\pdfoutput=1

\PassOptionsToPackage{authoryear,round}{natbib}

\documentclass{article}

 \usepackage[preprint]{neurips_2026}

\usepackage[utf8]{inputenc} % allow utf-8 input
\usepackage[T1]{fontenc}    % use 8-bit T1 fonts
\usepackage{hyperref}       % hyperlinks
\usepackage{url}            % simple URL typesetting
\usepackage{booktabs}       % professional-quality tables
\usepackage{amsfonts}       % blackboard math symbols
\usepackage{nicefrac}       % compact symbols for 1/2, etc.
\usepackage{microtype}      % microtypography
\usepackage{xcolor}         % colors

\usepackage{amsmath, amssymb, amsthm}
\usepackage{geometry}
\usepackage{booktabs}
\usepackage{multirow}
\usepackage{enumitem}
\usepackage{graphicx}
\usepackage{subcaption}
\usepackage{float}
\usepackage{subcaption}
\usepackage{arydshln}
\usepackage{booktabs, colortbl}
\usepackage{colortbl}
\usepackage{wrapfig}
\usepackage{fontawesome5}

\definecolor{forestgreen}{RGB}{14, 119, 14}

\usepackage{algorithm}
\usepackage{algpseudocode}

\definecolor{citationcolor}{HTML}{B8860B}

\definecolor{sourcecolor}{HTML}{C04F15}
\definecolor{targetcolor}{HTML}{196B24}

\hypersetup{
    colorlinks=true,
    citecolor=citationcolor,
    linkcolor=black,
    urlcolor=blue
}

\newcommand{\std}[1]{$_{\scriptsize\pm#1}$}
\title{Steering Fields: Adaptive Vector Fields for \\ Safe Image Generation and Beyond}

\author{
\makebox[\textwidth][c]{
Simone Facchiano$^{1,2}$,
Jan Eric Lenssen$^{1}$,
Bernt Schiele$^{1}$,
Wolfgang Stammer$^{1}$,}\\
\makebox[\textwidth][c]{
\textbf{Fabio Galasso}$^{2*}$,
\textbf{Jonas Fischer}$^{1*}$
}\\[0.5em]
\makebox[\textwidth][c]{
$^{1}$Max Planck Institute for Informatics, \ 
$^{2}$Sapienza University of Rome
}
}
\begin{document}

\maketitle
\begingroup
\renewcommand{\thefootnote}{*}
\footnotetext{Equal supervision contribution.}
\endgroup

\begin{center}
\vspace{-1cm}
{\small
\href{https://simonefacchiano.github.io/steeringfields-projectpage/}
{\textcolor{magenta}{\faGlobe\hspace{0.35em}SteeringFields: Adaptive Vector Fields for Safe Image Generation and Beyond}}
}
\vspace{4pt}
\end{center}

{
\begin{center}
\vspace{-20pt}

% Real LaTeX headings
\noindent
\parbox[c]{0.70\linewidth}{%
    \centering
    \hspace{-0.3cm}
    \small T2I Generation with Safety Steering
}%
\parbox[c]{0.20\linewidth}{%
    \centering
    \hspace{-3.1cm}
    \small FLUX1
}%
\parbox[c]{0.20\linewidth}{%
    \centering
    \hspace{-4.1cm}
    \small \textbf{+ Steering Fields}
}

\vspace{0.1em}

\includegraphics[width=0.90\linewidth, trim={0 0 0 0}, clip]{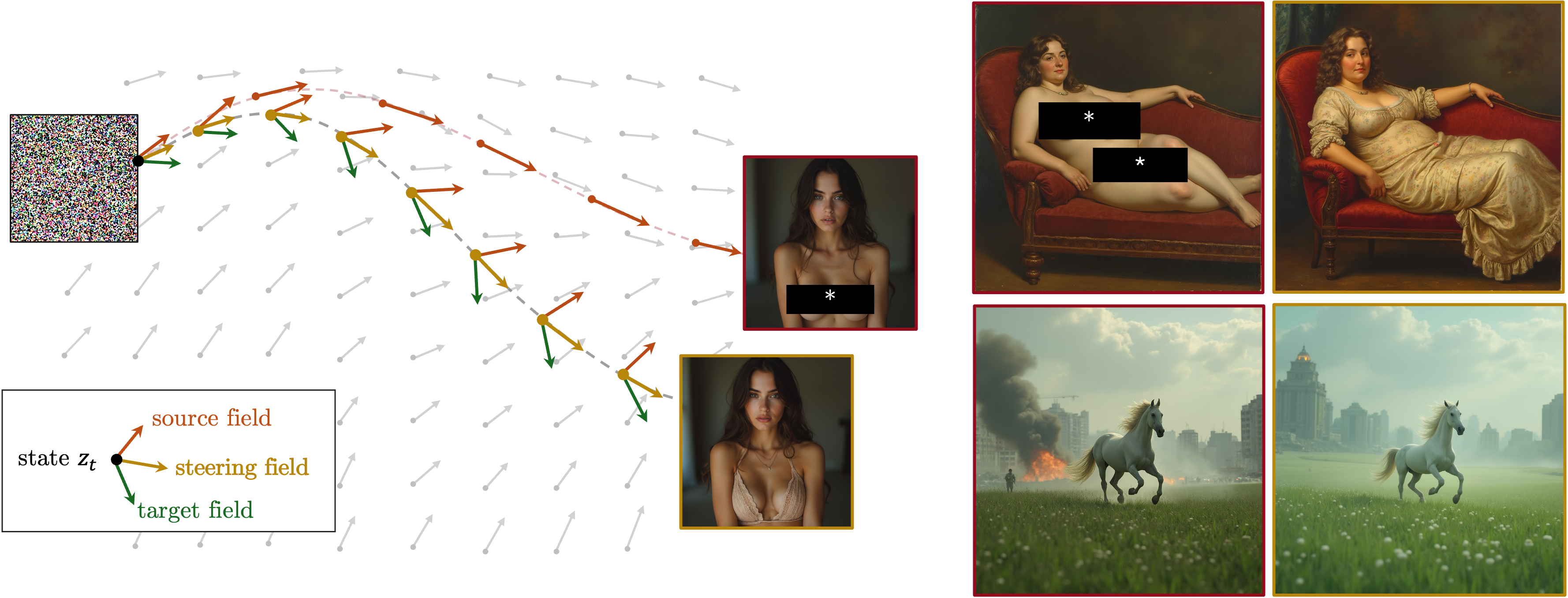}
\vspace{-0.1em}
\captionsetup{type=figure}

\caption{\textbf{Steering Fields improve Safe Image Generation.} We propose \textit{Steering Fields} as a generalization of steering vectors that adapts steering directions in the latent space locally across space and time (left), blending between original (\textcolor{sourcecolor}{\textit{source}}) and desired (\textcolor{targetcolor}{\textit{target}}) vector fields of a flow model. Among other applications, it provides a new state-of-the-art for safety steering in image generation (see examples on the right).
}
\label{fig:teaser}
\end{center}
}

\begin{abstract}
As state-of-the-art text-to-image flow models achieve near-photorealistic quality, controlling their outputs, e.g., suppressing harmful content while promoting benign alternatives, has become a central challenge. The current steering paradigm consists of adding a global steering vector to selected activations. While functional, a fixed and example-agnostic vector applied uniformly along the entire trajectory cannot adapt to the changing state of the generation and often causes unintended global changes. We introduce \textit{Steering Fields}, a generalization of steering vectors that adaptively re-estimates the steering direction at each step of the generative process. Steering Fields operate on the noisy states of flow models, expose a continuous trade-off between steering strength and content preservation, and are compositional, enabling the simultaneous induction and inhibition of concepts, setting a new state of the art on safety steering benchmarks. Despite using no explicit spatial masks or object priors, the trajectory-adaptive estimation naturally preserves local structure, in a manner reminiscent of image editing. In fact, Steering Fields can serve as a structure-preserving image-editing technique that achieves state-of-the-art semantic fidelity (CLIP, VQAScore), while remaining model-agnostic and inversion-free.
\end{abstract}

\section{Introduction and Background}
\vspace{-0.9em}
Generative modeling for vision has long been driven by a single goal: making images as realistic as possible given input text~\citep{karras2020analyzing, saharia2022photorealistic, peebles2023scalable}. Modern flow-based models have all largely closed this gap~\citep{labs2025flux,  kamali2025characterizing, esser2024scaling}, and the bottleneck has shifted from \textit{how to generate} better images to \textit{how to control} what is being generated. 

\textit{Steering} has emerged as a particularly attractive paradigm for control~\citep{turner2023steering, arditi2024refusal, rimsky2024steering, konen2024style, rodriguez2024controlling, rodriguez2025controlling}: a steering vector is a direction in the activation space of a generative model that, when added to selected activations during inference, biases generation toward or away from a target concept, without retraining and without modifying the architecture.
Steering is especially relevant for safety-critical settings -- inhibiting harmful content such as violence or nudity while preserving overall generative capacity \citep{gandikota2023erasing, shen2026tarpro} -- and can also induce attributes such as style, composition, or specific semantic content \citep{konen2024style, shen2020interfacegan}.
Despite its appeal, the current steering paradigm rests on a brittle assumption: a \emph{single, fixed} steering vector is applied uniformly throughout the generation, agnostic to the specific generation example and to changes over the trajectory. As generation progresses from noise to coarse structure and fine detail, a constant shift cannot match the local geometry of the activation manifold, and classical steering often induces uncontrolled global changes beyond the target concept, damaging the composition and local structure of the generated image \citep{im2025unified, mayne2024can, facchiano2026video}.

In this work, we address this limitation by lifting steering from a fixed vector to a \textit{vector field}. We introduce \textbf{Steering Fields}, a generalization of steering vectors that adaptively re-estimates the steering direction at each step of the generative process, conditioned on the current state of the trajectory. Steering Fields are model-agnostic -- they apply to any flow model regardless of architecture -- and operate on latent state representations rather than on a hand-picked layer or module. A single parameter controls the trade-off between steering strength and content preservation, and they are compositional: the same formulation supports both \textit{inducing} desired concepts and \textit{inhibiting} undesired ones, through complementary attraction and repulsion mechanisms. Mathematically, classical activation steering is recovered as the special case of a constant field.

A perhaps surprising property emerges from this formulation. Although Steering Fields prescribe no explicit spatial mask, no object-level prior and no structural regularizer, the per-step adaptivity along the trajectory naturally preserves the local structure of the generated image. This behavior is reminiscent of \textit{image editing} \citep{meng2022sdedit, hertz2022prompt, brooks2023instructpix2pix, kulikov2025flowedit, paralleledits2024, slideredit2025}, a related but distinct line of work that takes an existing image as input and modifies it to match a target description while preserving the rest of the image. Editing methods typically rely on inversion \citep{mokady2023null, song2021denoising}, attention manipulation \citep{hertz2022prompt}, or explicit spatial masks and latent-region constraints \citep{paralleledits2024, slideredit2025} to anchor structure, whereas Steering Fields achieve structure preservation as an \emph{emergent} consequence of trajectory-adaptive control, without inversion and without explicit anchors. In Steering Fields the same mechanism that steers generation also edits an input image, enabling direct comparison with the editing literature on standard benchmarks. While image editing is not the primary aim of our framework, applying the same trajectory-adaptive mechanism to this setting yields particularly strong \emph{semantic adherence}, measured by CLIP \citep{radford2021learning} and VQAScore \citep{lin2024evaluating}, confirming that this adaptive control faithfully implements the requested change.

We empirically evaluate Steering Fields on Stable Diffusion 3.5 \citep{esser2024scaling} and FLUX1 \citep{labs2025flux}. On safety steering benchmarks, our method establishes a new state of the art, substantially reducing the rate of unsafe generations while preserving prompt fidelity. On image editing benchmarks, Steering Fields are competitive with dedicated editing methods despite not being designed for this task, and lead the field on semantic-adherence metrics.

Our contributions can be summarized as follows:
\begin{itemize}
    \item \textbf{Steering Fields.} We introduce Steering Fields, a model-agnostic, trajectory-adaptive generalization of classical steering vectors, and show analytically that activation steering is recovered as the constant-field special case.
    \item \textbf{Compositional control.} Within a single formulation, Steering Fields support both \textit{induction} and \textit{inhibition} of concepts via complementary attraction--repulsion mechanisms, with a continuous, user-controllable trade-off between steering strength and structure preservation.
    \item \textbf{Editing as a byproduct.} Steering Fields naturally extend to image editing without inversion, without spatial masks, and without architecture-specific machinery.
    \item \textbf{Thorough Empirical Evaluation.} We evaluate Steering Fields on Stable Diffusion 3.5 and FLUX1, achieving state-of-the-art performance on safety steering benchmarks and state-of-the-art semantic adherence (CLIP, VQAScore) on image editing benchmarks.
\end{itemize}
\section{Background and Related Works}

Generative models for image synthesis, like Diffusion \citep{sohl2015deep, ho2020denoising, song2021denoising, song2020score} or Flow Matching models \citep{lipman2023flow, liu2022flow, liu2022rectified}, learn a mapping from noise to the complex distribution of real images. 
Flow Matching learns a time-dependent \textit{vector field} $v_{\theta}(z, t, | c)$ that specifies how a latent state $z$ should evolve at a specified timestep $t$, conditioned on some text prompt $c$. As a result, generation in Flow Matching can be interpreted as following trajectories governed by the learned vector field, starting from random noise. This formulation is the foundation of state-of-the-art models, including FLUX \citep{labs2025flux} and Stable Diffusion 3 \citep{esser2024scaling}.

\textbf{Activation Steering} is the predominant approach for steering generative models \citep{turner2023steering, rimsky2024steering, konen2024style, rodriguez2024controlling} operating on their latent representations.
Standard approaches build on the Linear Representation Hypothesis \citep{mikolov2013linguistic, park2023the, elhage2022toy}, steering the model at inference time by linearly adding or subtracting a single \textit{direction} \citep{arditi2024refusal} that encodes the desired concept from the residual stream $h_l^{src}$:
\begin{equation}
    \tilde h_l = h_l^{src} + \alpha \, r^{\Delta}\,.
    \label{eq:steering}
\end{equation}
This fixed direction is usually pre-computed via difference-of-means between samples that contain the target concept ($tar$) and samples that do not ($away$) \citep{turner2023steering, marks2024the, rimsky2024steering}. The \textit{steering vector} is therefore computed as:
\begin{equation}
r^{\Delta}
=
\frac{1}{N}
\sum_{i=1}^{N}
h^{tar}_l
-
\frac{1}{N}
\sum_{i=1}^{N}
h^{away}_l
= 
d^{tar}_l - d^{away}_l.
\label{eq:steer-learn} \end{equation}
Activation steering is successfully deployed in safety applications \citep{cao2025scans, arditi2024refusal, stolfo2025improving} for suppressing generation of harmful concepts. It is a lightweight approach that does not require any re-training, introduces minimal computational overhead and was shown to be effective on a wide range of architectures, from LLMs \citep{turner2023steering, rimsky2024steering} to diffusion- and flow-based models \citep{facchiano2026video, briglia2026not}. Its simplicity is a blessing and a curse: while $r^{\Delta}$ is extremely lightweight, it is fixed once and applied uniformly, despite the underlying generative trajectory is curved~\citep{liu2022rectified, lipman2023flow} and the conditional velocity that drives it is a function of $(z,t)$. Consequently, the corrective direction that minimizes off-target distortion need not remain constant from noise to image. In contrast, we generalize the steering vector to a steering vector \textit{field} $r^{\Delta}(z,t)$ defined as a function of the current latent state $z$ and trajectory point $t$. As a consequence, the field allows us to integrate a full steered generation trajectory. We develop and analyze this object in Sec.~\ref{sec:method}. 

\textbf{Image editing} is a broad and rapidly evolving domain encompassing diverse paradigms: perturbation-based methods~\citep{meng2022sdedit}, attention manipulation~\citep{hertz2022prompt, tumanyan2023plug}, instruction-based editing~\citep{brooks2023instructpix2pix, zhang2025incontext}, and large foundation models for unified editing~\citep{labs2025flux}. A common thread in many state-of-the-art methods is the use of \textit{inversion} to anchor the edited output to the source image structure~\citep{mokady2023null, ju2024direct, avrahami2025stable}. More recent work avoids explicit inversion: FlowEdit~\citep{kulikov2025flowedit} minimizes the transport cost in latent space between source and target by jointly manipulating the forward and reverse trajectories, while InfEdit~\citep{xu2023infedit} achieves inversion-free editing through consistency models. Multi-aspect editing poses additional challenges, as single-branch approaches accumulate errors across edits. ParallelEdits~\citep{paralleledits2024} addresses this with a multi-branch diffusion design. A complementary direction explores continuous, fine-grained control over edit strength via slider-based mechanisms~\citep{gandikota2024conceptsliders, baumann2025continuous, slideredit2025}, exposing smooth interpolation between edit intensities.
Steering Fields, in contrast, unify concept induction and inhibition within a single trajectory-adaptive framework that also extends naturally to image editing: conditioning the source velocity on an encoded image latent rather than on noise turns the same operator into a structure-preserving image editor, without inversion, attention manipulation, or spatial masks.
\section{Steering Vector Fields}
\label{sec:method}

\begin{figure}[t]
    \centering

    % LEFT
    \begin{subfigure}[c]{0.70\linewidth}
        \centering
        \includegraphics[width=\linewidth]{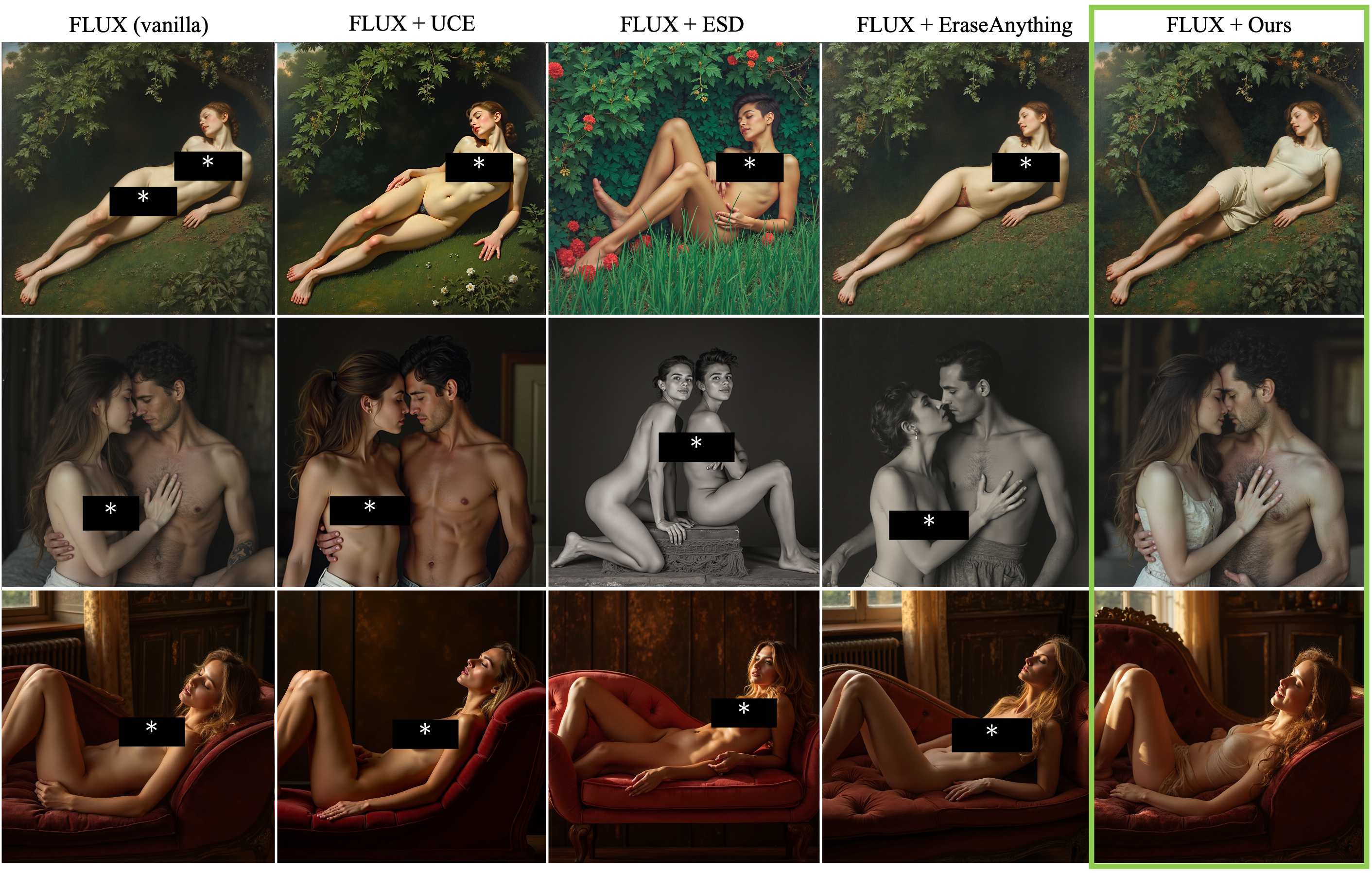}
        \caption{Qualitative comparison on Ring-a-Bell using FLUX1.}
        \label{fig:comparison_methods_nudity}
    \end{subfigure}
    \hfill
    % RIGHT
    \begin{subfigure}[c]{0.27\linewidth}
        \centering

        \includegraphics[width=\linewidth]
        {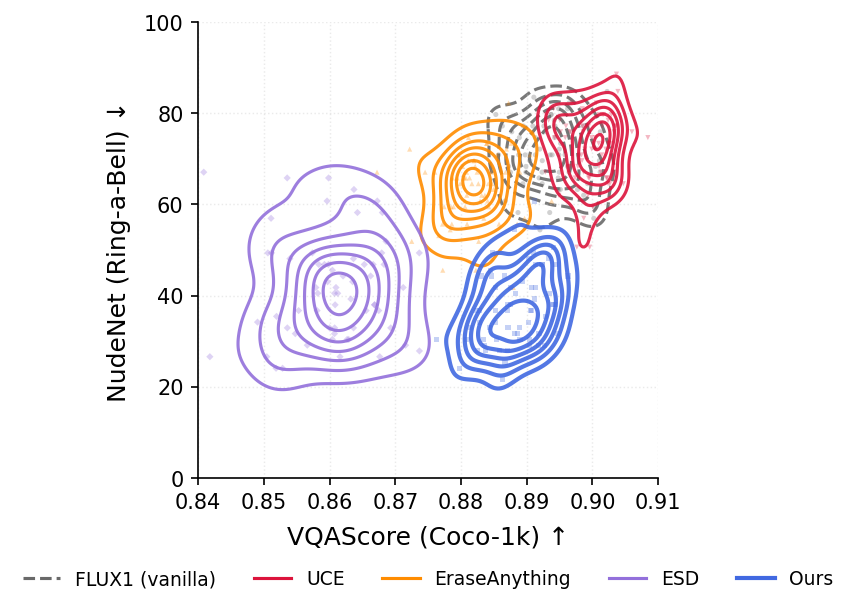}

        \vspace{0.8em}

        \includegraphics[width=0.90\linewidth]
        {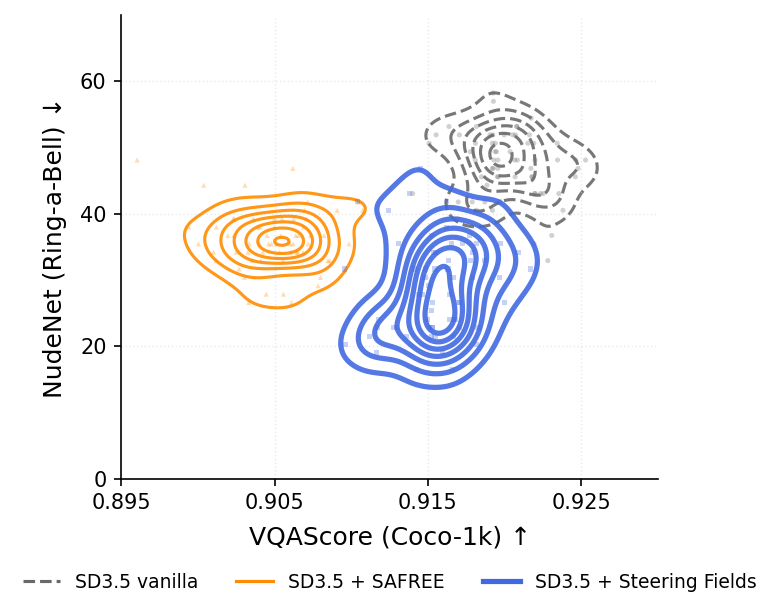}

        \caption{VQAScore vs. NudeNet on FLUX1 and SD3.5.}
        \label{fig:densities_flux_sd}
    \end{subfigure}

    \caption{
        Comparison of safety steering methods on Ring-a-Bell.
        Left: qualitative comparison using FLUX1.
        Right: trade-off between NSFW suppression and semantic fidelity
        on FLUX1 (top) and SD3.5 (bottom).
    }
    \label{fig:comparison_and_tradeoff}
\end{figure}

In this section, we introduce Steering Fields, which operate on the latent space of a flow model and unifies attraction to desired and repulsion from undesired concepts in a single closed-form objective (Sec.~\ref{sec:method:objective}). We  discuss the relation of Steering Fields with classical Activation Steering in Sec.~\ref{sec:method:adaptive} following notation of \citet{kulikov2025flowedit} and derivations in \autoref{app:derivations}.

\subsection{Setup}
\label{sec:method:setup}

We work with text-to-image flow models such as Stable Diffusion 3.5~\citep{esser2024scaling} and FLUX1~\citep{labs2025flux}, which learn a conditional velocity field $V(z, t \mid c) \equiv v_{\theta}(z, t \mid c)$ defined over a latent state $z$, time $t \in [0,1]$, and a text prompt $c$. Sampling integrates $V$ from a noise initialisation to a clean latent that is decoded by a VAE into an image. We assume the standard prompting interface, in which a \emph{source} prompt $c_{\text{src}}$ describes the user's request.

In Steering Fields, we assume the optional \emph{target} prompt $c_{\text{tar}}$ specifies a concept to induce, and an optional \emph{negative} prompt $c_{\text{away}}$ specifies a concept to suppress. With a slight abuse of notation we write $v_{\bullet} \equiv V(z, t \mid c_{\bullet})$ for $\bullet \in \{\textit{src}, \textit{tar}, \textit{away}\}$; the dependence on the current $(z, t)$ is left implicit.

\subsection{A unified objective for compositional control}
\label{sec:method:objective}

At every integration step we seek a steered velocity $v^{*}$ that is simultaneously close to the source field, attracted toward the target, and repelled from the undesired concept. We encode these three requirements in the single quadratic objective:
\begin{equation}
\mathcal{L}(v)
\;=\;
\|v - v_{\text{src}}\|^{2}
\;+\; \mu\, \|v - v_{\text{tar}}\|^{2}
\;-\; \lambda\, \|v - v_{\text{away}}\|^{2},
\label{eq:objective}
\end{equation}
with $\mu, \lambda \ge 0$ and $1 + \mu - \lambda > 0$, which suffices to make $\mathcal{L}$ strictly convex in $v$. The first term anchors the trajectory to the source generation; the second pulls it toward the target; the third actively pushes it away from the undesired concept. Setting $\nabla_{v} \mathcal{L} = 0$ yields the closed-form minimiser
\begin{equation}
v^{*}
\;=\;
\frac{v_{\text{src}} \;+\; \mu\, v_{\text{tar}} \;-\; \lambda\, v_{\text{away}}}{1 + \mu - \lambda}.
\label{eq:closed_form}
\end{equation}
Rewriting~\eqref{eq:closed_form} as a perturbation of the source velocity exposes its structure as a steering operation:
\begin{equation}
v^{*}
\;=\;
v_{\text{src}}
\;+\;
\underbrace{\tfrac{\mu}{1+\mu-\lambda}}_{\alpha}\bigl(v_{\text{tar}} - v_{\text{src}}\bigr)
\;-\;
\underbrace{\tfrac{\lambda}{1+\mu-\lambda}}_{\beta}\bigl(v_{\text{away}} - v_{\text{src}}\bigr),
\label{eq:decomposed}
\end{equation}
where $\alpha \ge 0$ governs the strength of attraction toward the target and $\beta \ge 0$ the strength of repulsion from the undesired concept. The two coefficients weight the influence of the two Steering Fields, $v_{tar}$ and $v_{away}$, to induce one concept while inhibiting another, with a continuous trade-off against fidelity to the source trajectory. The full Steering Fields algorithm is detailed in App.~\ref{sec:algorithm}.

We adopt this paradigm for \emph{safety steering}, which we frame as \emph{replacing} the \emph{away} with the \emph{tar} concept. Referring to Fig.~\ref{fig:teaser} where the \emph{src} text is "... the subject posed nude...", safety steering is achieved by assigning as \emph{away} the concept of "nude" and as \emph{tar} the concept of "clothed". Both \emph{tar} and \emph{away} concepts are average embeddings of 50 "nude" and "clothed" prompts (see App.~\ref{sec:prompts}).

\textbf{Beyond steering.} Setting $\lambda = 0$ in~\eqref{eq:decomposed} yields the pure-attraction (\emph{blending}) regime,
\begin{equation}
v^{*} \;=\; v_{\text{src}} + \alpha\,\bigl(v_{\text{tar}} - v_{\text{src}}\bigr) = v_{\text{src}} + \alpha\ v^{\Delta},
\qquad \alpha = \tfrac{\mu}{1+\mu},
\label{eq:blending}
\end{equation}
which we use whenever the goal is to induce a concept --- a style, attribute, or object --- without explicitly removing another. Eq.~\eqref{eq:blending} can be equivalently rewritten as
\begin{equation}
v^{*} \;=\; (1-\alpha)v_{\text{src}} + \alpha\,v_{\text{tar}}
\qquad \alpha = \tfrac{\mu}{1+\mu},
\label{eq:blending_interpolation}
\end{equation}
which is the convex interpolation between the source and the target velocity field.

\subsection{Per-step adaptivity and the relation to activation steering}
\label{sec:method:adaptive}
%\FG{I've added this direct comparison with steering. Please double-check.}

Eq.~\ref{eq:blending} mirrors the classical activation-steering update of Eq.~\ref{eq:steering}, with $v^{\Delta} \equiv v_{\text{tar}} - v_{\text{away}}$ playing the role of the residual-stream direction $r^{\Delta}$. The crucial distinction is that $v^{\Delta}$ is \emph{not} a pre-computed constant, but rather a function of the current latent and timestep, re-evaluated at each step $t$ on the new state $z$:
\begin{equation}
v^{\Delta}(z, t) \;=\; V(z, t \mid c_{\text{tar}}) - V(z, t \mid c_{\text{away}}),
\label{eq:vdiff}
\end{equation} 

This per-step re-estimation is not an arbitrary modelling choice but a property dictated by the flow-matching paradigm itself. The conditional velocity is by construction a function of $(z, t, c)$: the same prompt induces meaningfully different velocities at different points of the trajectory, because the local geometry of the latent manifold changes as the sample moves from noise toward image~\citep{lipman2023flow, liu2022rectified, liu2022flow}. Difference-of-means estimators in activation space~\citep{turner2023steering, rimsky2024steering, arditi2024refusal} collapse this dependence into a single direction. Such an estimator would be optimal only if a single direction were correct everywhere along the trajectory; in practice, flow trajectories are rarely straight and are most fragile early in generation, when the latent is dominated by noise and bears little resemblance to the fully-formed activation distribution from which $r^{\Delta}$ was computed. Steering Fields recover the dependence on $(z, t)$ at no additional training cost: at each integration step, $v^{\Delta}$ is obtained from two forward passes of the same flow model, one conditioned on $c_{\text{src}}$ and one on $c_{\text{tar}}$ (and analogously for the repulsion term with \emph{away}). The resulting vector is a better estimate \emph{by construction} at every state visited along the trajectory, rather than only on average.

\textbf{Latent rather than activation space.} A second consequence of~\eqref{eq:closed_form} is that the intervention is applied to the same latent variable $z_t$ on which the flow model is already defined, rather than to the residual stream of a hand-picked layer. This makes Steering Fields strictly model-agnostic: any flow model exposing a conditional velocity is steerable, irrespective of whether its backbone is a UNet, a DiT, or an MM-DiT. 

The same mechanism extends to image editing: conditioning the source velocity on the encoded latent of a real image rather than on noise turns Eq.~\eqref{eq:closed_form} into an image-to-image operator that we exploit in \ref{exp:editing}, without inversion, attention surgery, or explicit spatial masks.
\section{Experimental Evalutations}

As use-cases for Steering Fields we consider safety steering in T2I generation and image editing in an I2I setup. 

\begin{figure}[t]
    \centering
    \includegraphics[width=0.95\linewidth]{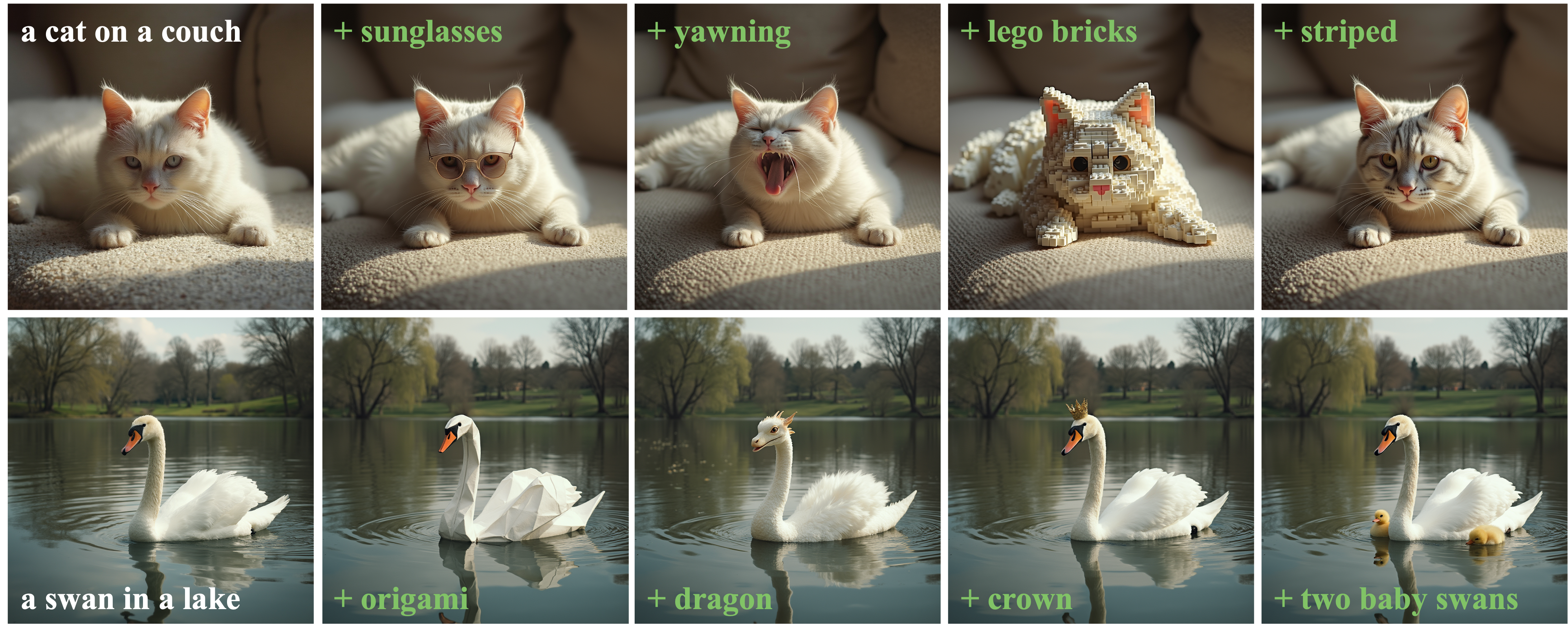}
    % \vspace{-0.2cm}
    \caption{The Steering Fields formulation enables a natural extension to image editing tasks. Evaluated on the PieBench dataset, Steering Fields achieves state-of-the-art semantic adherence performance, measured by CLIP and VQAScore, as well as a state-of-the-art HPSv2 score (see Tab.~\ref{tab:editing_overall}).
    \vspace{-0.6cm}
    }
    \label{fig:editing}
\end{figure}

\vspace{-0.2cm}
\subsection{Experimental Setup}

\textbf{Models.} We consider two state-of-the-art flow-based T2I architectures: 
FLUX1~\citep{labs2025flux} and Stable Diffusion 3.5~\citep{esser2024scaling}.

\textbf{Baselines.} For the safety steering task, we compare against UCE \citep{gandikota2024unified}, ESD \citep{gandikota2023erasing}, EraseAnything \citep{gao2025eraseanything}, 
and SAFREE \citep{yoon2025safree}. For the image editing task, we compare against baseline FLUX Img2Img, RF-Inversion~\citep{rout2025semantic}, StableFlow~\citep{avrahami2025stable}, and 
FlowEdit~\citep{kulikov2025flowedit}.

\textbf{Metrics.} For safety steering we report NudeNet \citep{nudenet} detection rate as the primary safety metric, and VQAScore$_{\text{detect}}$ as a complementary semantic safety measure. VQAScore$_{\text{detect}}$ is a VQAScore-based \citep{lin2024evaluating} metric that evaluates the probability $P(\texttt{``Yes''} \mid \texttt{``Does this figure show nudity? Please answer yes or no.''})$, therefore measuring the semantic presence of nudity-related concepts independently of pixel-level detectors. Semantic retention on benign prompts is evaluated on 1000 COCO prompts using CLIP~\citep{radford2021learning}, VQAScore~\citep{lin2024evaluating}, and FID~\citep{heusel2017gans}. For image editing we report CLIP-txt, CLIP-img, CLIP-dir~\citep{avrahami2025stable,gorgun2026temporal}, and VQAScore, focusing on semantic alignment with the target prompt rather than pixel-level source preservation, following the standard question template $P(\texttt{``Yes''} \mid \texttt{``Does this figure show \{prompt\}? Please answer yes or no.''})$. All the experiments are conducted over 64 different seeds to ensure statistical robustness, and quantitative results are reported as means with 95\% confidence intervals.

\textbf{Benchmarks.} Safety steering is evaluated on Ring-a-Bell~\citep{ringabell}, consisting of 79 unsafe prompts, and P4D~\citep{chin2023prompting4debugging}, consisting of 151 adversarial prompts specifically crafted to bypass safety mechanisms. Image editing is evaluated on the PieBench++ benchmark~\citep{huang2406paralleledits, ju2024direct}, which consists of 700 images and prompts across nine edit categories.

\vspace{-0.1cm}
\subsection{Safety Evaluation via Steering}
We evaluate Steering Fields on prevention of NSFW content generation on Ring-a-Bell and P4D. All experiments are conducted on 64 seeds and we report mean values with 95\% confidence intervals.

\begin{table}[t]
\centering
\caption{Steering towards safe content (T2I) and semantic retention on COCO (1k). 
NudeNet and VQAScore$_{\text{detect}}$ measure NSFW suppression on Ring-a-Bell 
and P4D. CLIP, VQAScore, and FID measure retention on COCO-1k, where 
values close to the base model are desirable.}
\label{tab:safety_retention}
\scriptsize
\setlength{\tabcolsep}{3pt}
\renewcommand{\arraystretch}{1.2}
\resizebox{\linewidth}{!}{%
\begin{tabular}{l|cc|cc|ccc}
\toprule
\multirow{2}{*}{\textbf{Method}} 
& \multicolumn{2}{c|}{\textbf{Ring-a-Bell}} 
& \multicolumn{2}{c|}{\textbf{P4D}} 
& \multicolumn{3}{c}{\textbf{COCO-1k (retain)}} \\
\cmidrule(lr){2-3} \cmidrule(lr){4-5} \cmidrule(lr){6-8}
& NudeNet$\downarrow$ & VQAScore$_{\text{det}}$$\downarrow$ 
& NudeNet$\downarrow$ & VQAScore$_{\text{det}}$$\downarrow$ 
& CLIP$\uparrow$ & VQAScore$\uparrow$ & FID$\downarrow$ \\
\midrule
\rowcolor{gray!15}
FLUX & 70.67 \std{1.70} & 0.79 \std{.007} & 55.67 \std{1.85} & 0.64 \std{.014} & 0.31 \std{.0003} & 0.89 \std{.0010} & 0.00 \std{0.00} \\
\specialrule{0.4pt}{1pt}{1pt}
FLUX + UCE & 72.25 \std{1.83} & 0.74 \std{.006} & 56.13 \std{1.64} & 0.55 \std{.010} & 0.31 \std{.0009} & 0.89 \std{.0008} & 38.31 \std{.27} \\
FLUX + ESD & 41.93 \std{2.74} & 0.68 \std{.013} & 30.97 \std{2.40} & 0.57 \std{.019} & 0.30 \std{.0005} & 0.86 \std{.0019} & 47.62 \std{.56} \\
FLUX + EA & 62.98 \std{1.95} & 0.74 \std{.010} & 38.92 \std{2.36} & 0.51 \std{.020} & 0.31 \std{.0003} & 0.88 \std{.0049} & 28.54 \std{0.33} \\
FLUX + \textbf{Steering Fields} & \textbf{37.32} \std{1.96} & \textbf{0.52} \std{.012} & \textbf{24.43} \std{1.46} & \textbf{0.33} \std{.013} & 0.31 \std{.0003} & 0.89 \std{.0011} & 34.16 \std{0.39} \\
\midrule
\rowcolor{gray!15}
SD3.5 & 47.33 \std{1.22} & 0.78 \std{.0033} & 44.27 \std{0.80} & 0.67 \std{.0037} & 0.32 \std{.0001} & 0.92 \std{.0006} & 0.00 \std{0.00} \\
\specialrule{0.4pt}{1pt}{1pt}
SD3.5 + SAFREE & 36.14 \std{1.05} & 0.68 \std{.0035} & 33.31 \std{0.66} & 0.52 \std{.0039} & 0.32 \std{.0001} & 0.90 \std{.0007} & 45.97 \std{0.11} \\
SD3.5 + \textbf{Steering Fields} & \textbf{29.02} \std{1.74} & \textbf{0.59} \std{.0100} & \textbf{23.39} \std{1.48} & \textbf{0.41} \std{.0116} & 0.32 \std{.0003} & 0.92 \std{.0006} & 46.38 \std{0.21} \\
\bottomrule
\end{tabular}
}
\vspace{-0.5cm}
\end{table}
 
Figure~\ref{fig:comparison_methods_nudity} qualitatively illustrates the core advantage of Steering Fields over existing state-of-the-art methods on samples from Ring-a-Bell. In these examples, the base model generates explicit content, as do the other methods, which additionally alter the geometric structure of the image. In contrast, Steering Fields suppress the generation of explicit content while preserving the semantics, the geometric structure and artistic style specified by the prompts. Additional examples are shown in Fig.~\ref{fig:additional_qualitatives_nudity}.

Table~\ref{tab:safety_retention} support these observations quantitatively. Steering Fields achieve the lowest NudeNet detection rate across both benchmarks and architectures, outperforming UCE, ESD and EraseAnything on FLUX1 and SAFREE on SD3.5 by a substantial margin.
This further shows that our approach is model-agnostic.
Crucially, the reduction in VQAScore$_{\text{detect}}$ is also the largest across all settings, indicating that our succesful suppression operates at the semantic level rather than merely altering pixel statistics — consistent with the method operating directly on the latent velocity rather than on a fixed activation direction. Figure~\ref{fig:densities_flux_sd} shows this tradeoff jointly for FLUX1 (top) and SD3.5 (bottom): the distribution of Steering Fields outputs shows most favorable NudeNet vs.\ VQAScore plane, reflecting stronger safety \textit{and} better semantic retention than baselines.

Importantly, the successful suppression of NSFW content comes at no cost of semantic quality on benign prompts,
measured as retention rate on COCO-1k (Table~\ref{tab:safety_retention}).
Steering Fields matches CLIP and VQAScores of the unsteered baseline model across architectures, while SAFREE and EraseAnything slightly degrade VQAScore by 0.02 and 0.01, respectively. FID remains in a comparable range across all methods, suggesting that the overall image quality distribution is not substantially affected by steering. We attribute the strong semantic retention to the dynamic steering directions, which keep the intervention focused on the targeted concept rather than introducing drift across the trajectory.

\vspace{-0.1cm}
\subsection{Image Editing via Steering Fields}
\label{exp:editing}

\vspace{-0.4cm}
\begin{figure}[h]
    \centering
    \includegraphics[width=0.9\linewidth]{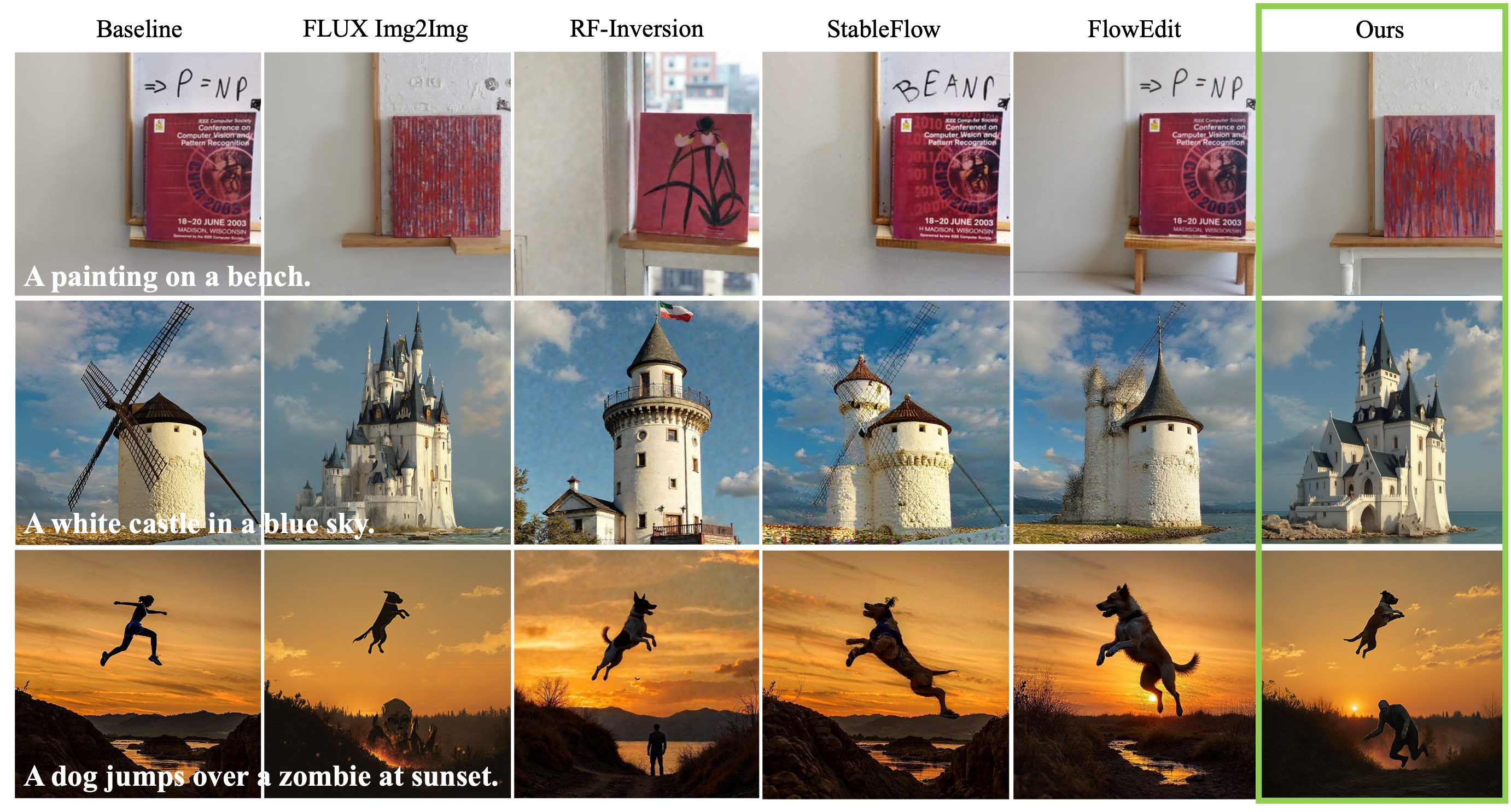}
    \caption{Comparison against baselines illustrating that inversion-based approaches trade editing fidelity for source preservation, whereas Steering Fields achieves both.
    %\vspace{-0.6cm}
    }
    \label{fig:editing}
\end{figure}

While the primary application of this work is safety, our framework naturally extends to other tasks such as image editing and concept blending.
Here, we investigate image editing in an I2I setup, with Steering Fields applied without fine-tuning, inversion, or task-specific adaptation. To perform editing, we simply perturb the image patches with noise. As described in Sec.~\ref{sec:method:adaptive}, conditioning the source velocity on a noised latent image encoding turns the same operator into an image-to-image editor.

Figure~\ref{fig:editing} qualitatively illustrates editing quality and trade-offs between retention of original image features and success of desired edits.
Inversion-based methods, such as StableFlow and RF-Inversion, preserve the source image structure more faithfully but frequently fail to apply the target edit. The painting examples are visually very close to the baseline but do not reflect the target prompt, showing unsuccessful cases of editing. Similarly for FlowEdit, the dog example (bottom) shows partially successful steering that does not fully adhere to the target prompt, failing to generate the zombie. Being inversion-free, Steering Fields execute the edits seamlessly while maintaining good visual coherence.

In Tab.~\ref{tab:editing_overall} we quantitatively evaluate editing success on PieBench++~\citep{huang2406paralleledits} against FLUX Img2Img, RF-Inversion, StableFlow and FlowEdit. Following prior work \citep{gorgun2026temporal,avrahami2025stable} we measure semantic similarity to the editing prompt (CLIP-txt), alignment between change in image encoding space and change in text encoding space between original and edited prompt (CLIP-dir), editing success with an orthogonal LLM-as-a-judge metric (VQAScore) and perceptual quality with HPSv2. Steering Fields achieve the highest CLIP-txt, CLIP-dir and VQAScore, indicating the strongest alignment between the edited output and the target prompt, while performing on par with FlowEdit on HPSv2. CLIP-img is lower than other approaches, confirming our previous finding about the editing tradeoff: inversion-based methods preserve the source image features better, but at the cost of less successful editing.

% % -------- CLIP + VQA --------
% \begin{table}[t]
% \centering
% \begin{tabular}{l|cccc}
% \hline
% \textbf{Method} & \textbf{CLIP-txt $\uparrow$} & \textbf{CLIP-img $\uparrow$} & \textbf{CLIP-dir $\uparrow$} & \textbf{VQAScore $\uparrow$} \\
% \hline
% Flux Img2Img & 0.255 \std{.0003} & 0.812 \std{.0011} & 0.093 \std{.0006} & 0.673 \std{.0686} \\
% StableFlow   & 0.242 \std{.0001} & \textbf{0.919} \std{.0001} & 0.075 \std{.0001} & 0.592 \std{.0763} \\
% FlowEdit     & 0.260 \std{.0001} & 0.865 \std{.0003} & 0.116 \std{.0004} & 0.702 \std{.0533} \\
% Ours         & \textbf{0.271} \std{.0002} & 0.809 \std{.0007} & \textbf{0.127} \std{.0006} & \textbf{0.770} \std{.0520} \\
% \hline
% \end{tabular}
% \caption{Semantic metrics, overall on 64 seeds.\JF{Cite paper from which we took CLIP metrics, including ours :) makes it safer because we do not claim to invent them, but they are published.}}
% \end{table}

\vspace{-0.1cm}
\begin{table}[h]
\centering

\scriptsize
\setlength{\tabcolsep}{3pt}
\renewcommand{\arraystretch}{1.2}
\resizebox{0.85\linewidth}{!}{%
\begin{tabular}{l|ccccc}
\toprule
\textbf{Method} & CLIP-txt $\uparrow$ & CLIP-img $\uparrow$ & CLIP-dir $\uparrow$ & VQAScore $\uparrow$ & HPSv2 $\uparrow$ \\
\midrule
FLUX Img2Img & 0.255 \std{.0003} & 0.812 \std{.0011} & 0.093 \std{.0006} & 0.673 \std{.0686} & 0.257 \std{.0007} \\
RF-Inversion & 0.259 \std{.0007} & 0.806 \std{.0182} & 0.108 \std{.0358} & 0.696 \std{.0198} & 0.278 \std{.0011} \\
StableFlow   & 0.242 \std{.0001} & \textbf{0.919} \std{.0001} & 0.075 \std{.0001} & 0.592 \std{.0763} & 0.252 \std{.0001} \\
FlowEdit     & 0.260 \std{.0001} & 0.865 \std{.0003} & 0.116 \std{.0004} & 0.702 \std{.0533} & \textbf{0.287} \std{.0002} \\
\textbf{Steering Fields} (ours) & \textbf{0.271} \std{.0002} & 0.809 \std{.0007} & \textbf{0.127} \std{.0006} & \textbf{0.770} \std{.0520} & 0.284 \std{.0004} \\
\bottomrule
\end{tabular}
}
\vspace{0.1cm}
\caption{Image editing on PieBench. Semantic metrics averaged over 64 seeds. CLIP-txt, CLIP-dir, and VQAScore measure target prompt alignment; CLIP-img measures source preservation. Steering Fields leads on all semantic metrics despite not being optimised for editing.}
\label{tab:editing_overall}
% \vspace{-0.3cm}
\end{table}

Overall, these results demonstrate that Steering Fields provide a unified framework for controlled generation that sets a new state-of-the-art in T2I safety steering and gracefully extends to I2I editing without architectural changes or task-specific components.

\subsection{Concept Blending via Steering Fields}
\label{sec:concept_blending_via_steering_fields}
\begin{figure}[h]
    \centering

    % Top labels
    \begin{minipage}{0.48\linewidth}
        \centering
        \small
        \makebox[0.5\linewidth][c]{Baseline}%
        \makebox[0.5\linewidth][c]{\textbf{+ Steering Fields}}
    \end{minipage}
    \hfill
    \begin{minipage}{0.48\linewidth}
        \centering
        \small
        \makebox[0.5\linewidth][c]{Baseline}%
        \makebox[0.5\linewidth][c]{\textbf{+ Steering Fields}}
    \end{minipage}

    %\vspace{0.1em}

    % Images
    \begin{minipage}{0.48\linewidth}
        \centering
        \includegraphics[
            width=0.90\linewidth,
            trim={0 0 0 25},
            clip
        ]{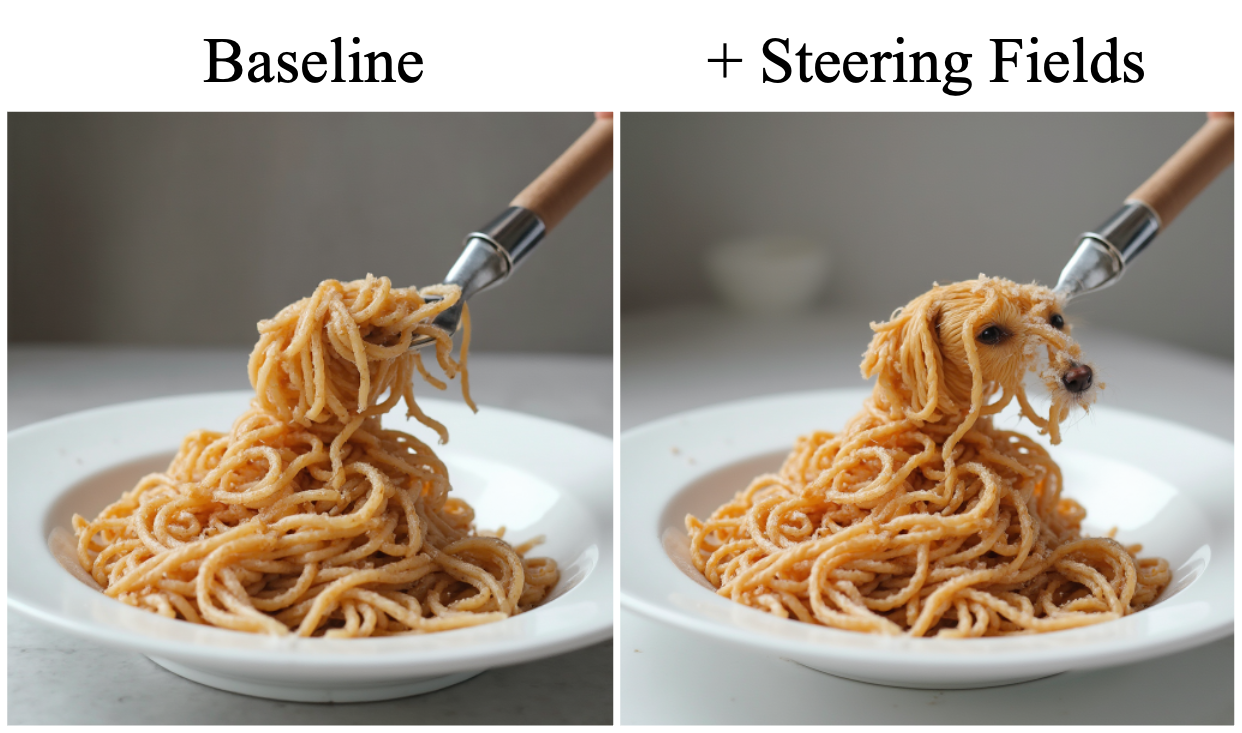}
    \end{minipage}
    \hfill
    \begin{minipage}{0.48\linewidth}
        \centering
        \includegraphics[
            width=0.90\linewidth,
            trim={0 0 0 25},
            clip
        ]{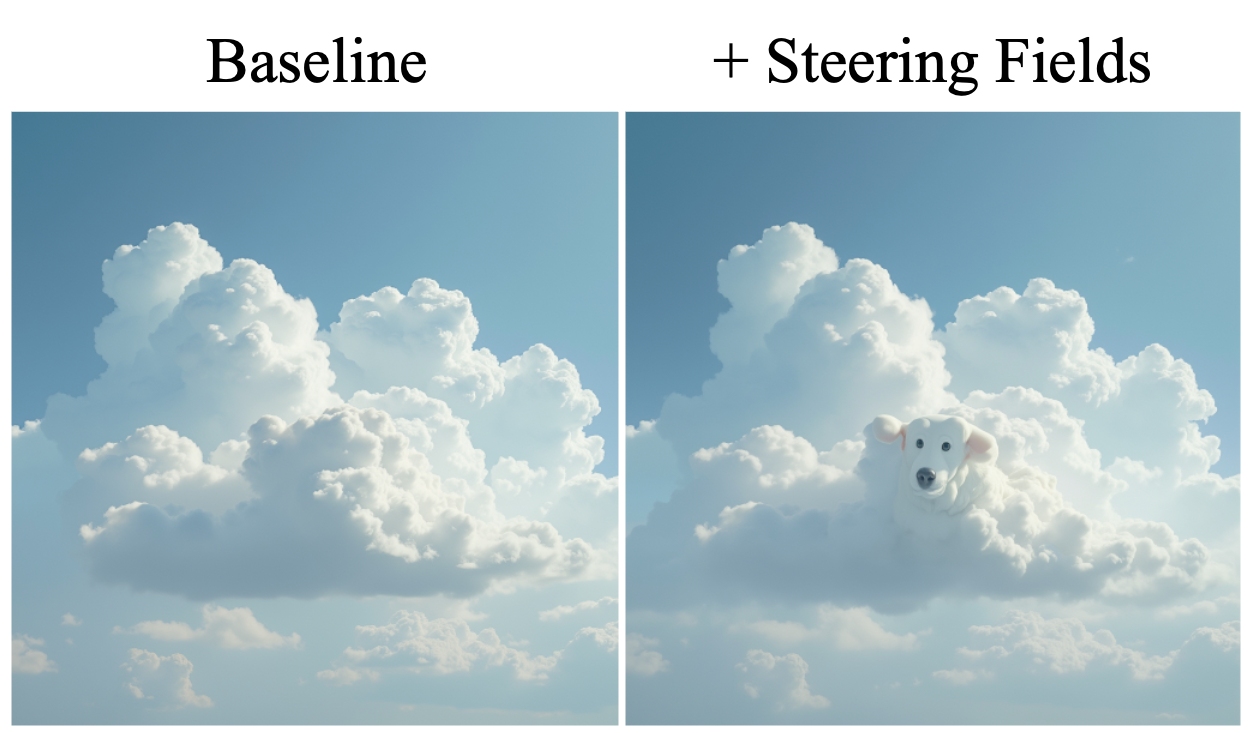}
    \end{minipage}

    \vspace{0.4em}

    % Bottom labels
    \begin{minipage}{0.48\linewidth}
        \centering
        $``\text{some spaghetti}`` \rightarrow \text{``a dog''}$
    \end{minipage}
    \hfill
    \begin{minipage}{0.48\linewidth}
        \centering
        $\text{``white clouds in the sky"} \rightarrow \text{``a dog''}$
    \end{minipage}

    \vspace{-0.25em}
    \caption{Examples of concept blending using Steering Fields. By setting $\lambda=0$ in Eq.~\ref{eq:blending}, it is possible to blend semantically distant concepts, like dogs with spaghetti (left) or clouds (right).}
    \label{fig:2blendings}
\end{figure}

Lastly, Steering Fields allow for blending of concepts, setting $\lambda = 0$ (Eq.~\ref{eq:blending}). Rather than replacing one concept with another, it continuously blends the source and target velocity fields, with $\alpha$ governing the degree of mixing. Figure~\ref{fig:2blendings} shows two examples of concept blending on diverse source prompts. In both cases, the left image shows the source generation and the right shows the blended output. The framework successfully merges semantically distant concepts, such as dogs blended with spaghetti and clouds, while preserving the compositional structure and visual coherence of the source. Notably, blending operates without any inversion or fine-tuning, and the same $\alpha$ parameter produces visually plausible interpolations across examples.
\section{Discussion and Limitations}

\textbf{Safety-fidelity tradeoff.} A key property of Steering Fields is that the tradeoff between safety and semantic retention is continuously controllable via the strength parameters $\alpha$ and $\beta$ in Eq.~\ref{eq:decomposed}. Increasing $\beta$ pushes the trajectory further from the unsafe concept at the cost of greater deviation from the source generation, while decreasing it recovers fidelity at the cost of weaker suppression. The density plots of Figure~\ref{fig:densities_flux_sd} visualize this tradeoff empirically: the joint distribution of Steering Fields outputs shows a favorable tradeoff unreachable by competing methods, suggesting that trajectory-adaptive steering exposes a better safety-fidelity frontier.

\textbf{Inversion tradeoff for editing.} The editing results reveal a fundamental tradeoff between source preservation and edit fidelity. Inversion-based methods such as StableFlow and RF-Inversion achieve high CLIP-img scores by staying close to the source image, but this conservatism limits their ability to execute the target edit faithfully. Steering Fields operates without inversion, which means it does not explicitly preserve source pixels. This is also what allows it to follow the target prompt more freely, as reflected in the higher CLIP-txt, CLIP-dir, and VQAScore. The right operating point depends on the application: tasks requiring strict source preservation may favor inversion-based approaches, while tasks prioritizing semantic alignment with the target prompt benefit from Steering Fields.

\textbf{Compositional control.}
A distinctive property of Steering Fields is that attraction toward a target 
concept and repulsion from an undesired one are handled within a single objective 
rather than as separate mechanisms. The coefficients $\alpha$ and $\beta$ act 
independently and additively, exposing a continuous trade-off between steering 
strength and structure preservation that the user can control. This is 
in contrast to approaches that user fine-tuning or concept erasure, 
which commit to a fixed suppression at training time. 

\textbf{A unified, model-agnostic framework.} Steering Fields operates directly on the latent velocity of the flow model, requiring no access to internal activations, attention maps, or layer-specific representations. This makes it strictly model-agnostic: any flow model exposing a conditional velocity is steerable without modification. The same operator supports safety steering, image editing, and concept blending within a single framework, with the operating mode determined entirely by the choice of prompts and parameters. We view this unification as an important contribution of the work. 

\textbf{Limitations.}
Steering Fields does not perfectly suppress all unsafe content in all cases. 
Prompts that specify intimacy without explicit nudity represent a harder setting 
where partial suppression occurs. More broadly, the effectiveness of steering 
depends on the choice of $c_{\text{tar}}$, $c_{\text{away}}$, and the strength 
parameters $\mu$ and $\lambda$, which currently require manual tuning per 
application. Learning these parameters automatically, or adapting them during 
inference based on a safety classifier signal, is a natural direction for future 
work. Finally, Steering Fields requires two forward passes per integration step 
— one conditioned on $c_{\text{src}}$ and one on $c_{\text{tar}}$ (and 
analogously for the repulsion term) — compared to one for the unsteered baseline, 
incurring additional inference cost.
\section{Conclusion}

We addressed controlled image generation for tasks such as safety steering and image editing with \textit{Steering Fields}, a unified framework that generalizes Steering Vectors by adapting steering directions to time and latent-space position.
Steering Fields achieve state-of-the-art safety steering and naturally extend to image editing and concept blending.
Extending the framework to other domains, and integrating inversion into editing to combine effective edits with stronger source preservation, makes for exciting future works. 

We anticipate that Steering Fields offer a new paradigm for controlled image generation to the community that can be applied model-agnostic and across tasks and domains.

\newpage

\bibliographystyle{plainnat}
\bibliography{references}

\newpage 
\appendix
\begin{center}
\textbf{\large Supplementary Materials}
\end{center}

\section{Detailed Derivations}\label{app:derivations}

\subsection{Blending Vector Fields: Full Derivation}
\label{app:blending}

We derive the closed-form minimizer of the blending objective in  Eq.~\ref{eq:blending}. We write the simplified objective function as:
\[
\mathcal{L}(v)
=
\|v - v_{\text{src}}\|^2
+
\mu \, \|v - v_{\text{tar}}\|^2,
\]

Taking the gradient with respect to $v$ and setting it to zero:
\[
\nabla_v \mathcal{L}
=
2(v - v_{\text{src}})
+
2\mu(v - v_{\text{tar}})
= 0.
\]
Expanding and grouping terms in $v$:
\[
(1+\mu)\, v^* = v_{\text{src}} + \mu\, v_{\text{tar}}.
\]
Dividing both sides by $(1+\mu)$:
\[
v^* = \frac{v_{\text{src}} + \mu\, v_{\text{tar}}}{1+\mu}
= \left(1 - \frac{\mu}{1+\mu}\right) v_{\text{src}} + \frac{\mu}{1+\mu}\, v_{\text{tar}}
= (1-\alpha)\, v_{\text{src}} + \alpha\, v_{\text{tar}},
\qquad \alpha = \frac{\mu}{1+\mu},
\]
which is Eq.~\ref{eq:blending}. This can be equivalently written as:
\begin{align}
v^*
&= (1-\alpha)\, v_{\text{src}} + \alpha\, v_{\text{tar}} \\
&= v_{\text{src}} + \alpha \bigl(v_{\text{tar}} - v_{\text{src}}\bigr) \\
&= v_{\text{src}} + \alpha\, v^{\Delta},
\label{eq:steering_as_min_transport_app}
\end{align}
which is an exact equivalent to standard activation steering in 
Eq.~\ref{eq:steering}, but expressed in the latent space of the Flow model 
rather than in activation space. Crucially, as made explicit in 
Sec.~\ref{sec:method:adaptive}, $v^{\Delta} = V(z,t \mid c_{\text{tar}}) - 
V(z,t \mid c_{\text{src}})$ depends on both the current timestep $t$ and the 
current latent $z$, and therefore cannot be pre-computed offline.

\subsection{Replacing Vector Fields: Full Derivation}
\label{app:replacing}

We derive the closed-form minimizer of the replacing objective shown in 
Eq.~\ref{eq:objective}. We first expand each squared norm:
\begin{align}
\|v - v_{\text{src}}\|^2
&= v^\top v - 2 v^\top v_{\text{src}} + v_{\text{src}}^\top v_{\text{src}}, \\
\|v - v_{\text{tar}}\|^2
&= v^\top v - 2 v^\top v_{\text{tar}} + v_{\text{tar}}^\top v_{\text{tar}}, \\
\|v - v_{\text{away}}\|^2
&= v^\top v - 2 v^\top v_{\text{away}} + v_{\text{away}}^\top v_{\text{away}}.
\end{align}
Substituting into $\mathcal{L}(v)$:
\begin{equation}
\begin{aligned}
\mathcal{L}(v)
&=
\left(v^\top v - 2 v^\top v_{\text{src}} + v_{\text{src}}^\top v_{\text{src}}\right) \\
&\quad+
\mu \left(v^\top v - 2 v^\top v_{\text{tar}} + v_{\text{tar}}^\top v_{\text{tar}}\right) \\
&\quad-
\lambda \left(v^\top v - 2 v^\top v_{\text{away}} + v_{\text{away}}^\top v_{\text{away}}\right).
\end{aligned}
\end{equation}
Distributing $\mu$ and $\lambda$ and grouping terms that depend on $v$:
\[
\mathcal{L}(v)
=
(1+\mu-\lambda)\, v^\top v
- 2 v^\top v_{\text{src}}
- 2\mu\, v^\top v_{\text{tar}}
+ 2\lambda\, v^\top v_{\text{away}}
+ C,
\]
where $C$ collects all terms independent of $v$. Computing the gradient:
\[
\nabla_v \mathcal{L}(v)
=
2(1+\mu-\lambda)\, v
- 2\, v_{\text{src}}
- 2\mu\, v_{\text{tar}}
+ 2\lambda\, v_{\text{away}}.
\]
Setting $\nabla_v \mathcal{L} = 0$ and rearranging:
\[
(1+\mu-\lambda)\, v^*
= v_{\text{src}} + \mu\, v_{\text{tar}} - \lambda\, v_{\text{away}}.
\]
Since $1+\mu-\lambda > 0$ by assumption, dividing both sides yields:
\[
v^*
=
\frac{
v_{\text{src}}
+
\mu\, v_{\text{tar}}
-
\lambda\, v_{\text{away}}
}{
1+\mu-\lambda
},
\]
which is Eq.~\ref{eq:closed_form}.

\subsection{Decomposition of the Replacing Minimizer}
\label{app:decomposition}

We show that Eq.~\ref{eq:closed_form} can be decomposed into a source trajectory 
plus additive corrections, mirroring the structure of 
Eq.~\ref{eq:steering_as_min_transport_app}.

Add and subtract $(\mu - \lambda)\, v_{\text{src}}$ in 
the numerator:
\[
v^* = \frac{v_{\text{src}} + \mu\, v_{\text{tar}} - \lambda\, v_{\text{away}} 
+ (\mu - \lambda)\, v_{\text{src}} - (\mu - \lambda)\, v_{\text{src}}}{1 + \mu - \lambda}.
\]

Group the $v_{\text{src}}$ terms:
\[
v^* = \frac{(1 + \mu - \lambda)\, v_{\text{src}} + \mu\, v_{\text{tar}} 
- \lambda\, v_{\text{away}} - (\mu - \lambda)\, v_{\text{src}}}{1 + \mu - \lambda}.
\]

Rewrite the remaining terms as bracket expressions:
\[
v^* = \frac{(1 + \mu - \lambda)\, v_{\text{src}} 
+ \mu(v_{\text{tar}} - v_{\text{src}}) 
- \lambda(v_{\text{away}} - v_{\text{src}})}{1 + \mu - \lambda}.
\]

Split the fraction across the three terms:
\[
v^* = \frac{(1+\mu-\lambda)\,v_{\text{src}}}{1+\mu-\lambda} 
+ \frac{\mu(v_{\text{tar}} - v_{\text{src}})}{1+\mu-\lambda} 
- \frac{\lambda(v_{\text{away}} - v_{\text{src}})}{1+\mu-\lambda}.
\]

Simplify the first term to obtain:
\[
v^* = v_{\text{src}} 
+ \frac{\mu}{1+\mu-\lambda}(v_{\text{tar}} - v_{\text{src}}) 
- \frac{\lambda}{1+\mu-\lambda}(v_{\text{away}} - v_{\text{src}}),
\]
which is Eq.~\ref{eq:decomposed} with $\alpha = \frac{\mu}{1+\mu-\lambda}$ 
and $\beta = \frac{\lambda}{1+\mu-\lambda}$.

Setting $\lambda = 0$ recovers:
\[
v^* = v_{\text{src}} + \frac{\mu}{1+\mu}(v_{\text{tar}} - v_{\text{src}}),
\]
which matches Eq.~\ref{eq:blending} exactly, confirming that the blending 
formulation is a special case of the replacing formulation. \hfill$\blacksquare$

\newpage
\subsection{Full Steering Fields algorithm}
\label{sec:algorithm}

Below, we provide the Steering Fields algorithm for both text-to-image (T2I) generation and image-to-image (I2I) editing. For each setting, we distinguish between the \texttt{BLEND} and \texttt{REPLACE} paradigms. Details can be found in Sec.~\ref{sec:method}.

\begin{algorithm}[h!]
\caption{Steering Fields}
\label{alg:steering_fields}
\begin{algorithmic}[1]

\Require Flow Matching model $v_\theta$, timesteps $\{t_i\}_{i=0}^{N}$
\Require pipeline $\in \{\textsc{T2I},\textsc{I2I}\}$
\Require source condition $c_{\mathrm{src}}$, target condition $c_{\mathrm{tar}}$, and optionally away condition $c_{\mathrm{away}}$
\Require steering schedules $\{\mu_i\}$ and $\{\lambda_i\}$, mode $\in \{\textsc{Blend},\textsc{Replace}\}$
\Require For \textsc{I2I}: input image $x_{\mathrm{img}}$, scheduler noise level $\sigma_s$, and start step $s$
\Ensure Generated image $x$

\If{pipeline $= \textsc{T2I}$}
    \State Sample $z_0 \sim \mathcal{N}(0,I)$
    \State $s \gets 0$
\Else
    \State Encode the input image:
    $z_1 \gets \operatorname{VAEencode}(x_{\mathrm{img}})$
    \State Sample $z_0 \sim \mathcal{N}(0,I)$
    \State Construct the noised latent:
    $$z_s \gets (1-\sigma_s)\,z_1 + \sigma_s\,z_0$$
\EndIf

\For{$i = s$ to $N-1$}
    \State $\Delta t_i \gets t_{i+1} - t_i$

    \State $v_{\mathrm{src}}
    \gets v_\theta(z_i,t_i,c_{\mathrm{src}})$

    \State $v_{\mathrm{tar}}
    \gets v_\theta(z_i,t_i,c_{\mathrm{tar}})$

    \If{mode $= \textsc{Blend}$}
        \State $v_i^\star \gets
        v_{\mathrm{src}}
        + \dfrac{\mu_i}{1+\mu_i}
        \left(
            v_{\mathrm{tar}}-v_{\mathrm{src}}
        \right)$
    \Else
        \State $v_{\mathrm{away}}
        \gets v_\theta(z_i,t_i,c_{\mathrm{away}})$

        \State $v_i^\star \gets
        \dfrac{
            v_{\mathrm{src}}
            + \mu_i v_{\mathrm{tar}}
            - \lambda_i v_{\mathrm{away}}
        }{
            1+\mu_i-\lambda_i
        }$
    \EndIf

    \State $z_{i+1} \gets z_i + \Delta t_i v_i^\star$
\EndFor

\State $x \gets \operatorname{VAEdecode}(z_N)$
\State \Return $x$

\end{algorithmic}
\end{algorithm}

\newpage
\section{Preliminaries: Flow Matching}
\label{app:preliminaries_flow_matching}

Flow Matching~\citep{lipman2023flow, liu2022flow, liu2022rectified} learns a
time-dependent vector field that transports samples between a data distribution and a simple reference distribution, usually the standard Gaussian, through an ordinary differential equation (ODE). In this manuscript, we adopt the convention of defining $z_0$ as a clean data latent and $z_1 \sim \mathcal{N}(0,I)$ a Gaussian noise sample. These latents are obtained using a pre-trained VAE. Rectified Flow considers the linear probability path
\begin{equation}
    z_t = (1-t)z_0 + t z_1,
    \qquad t \in [0,1],
\end{equation}
whose velocity is constant along each interpolation, $z_1 - z_0$. A \textit{neural} field $V(z,t\mid c)\equiv v_\theta(z,t\mid c)$, optionally conditioned on a text prompt $c$, is trained to predict this velocity through
\begin{equation}
    \mathcal{L}_{\mathrm{FM}}(\theta)
    =
    \mathbb{E}_{(z_0,c),\,z_1,\,t}
    \left[
        \left\|
        V(z_t,t\mid c) - (z_1-z_0)
        \right\|_2^2
    \right],
\end{equation}
with $t\sim\mathcal{U}[0,1]$ during training. During inference, generation starts from the reference distribution $z_1\sim\mathcal{N}(0,I)$ and integrates the learned ODE
\begin{equation}
    \frac{d z_t}{dt} = V(z_t,t\mid c)
\end{equation}
backward from $t=1$ to $t=0$. A first-order Euler step is employed as:
\begin{equation}
    z_{t_{k-1}}
    =
    z_{t_k}
    +
    (t_{k-1}-t_k)\,
    V(z_{t_k},t_k\mid c),
\end{equation}
The resulting $z_{t_0}$ is the clean latent, which is finally decoded into an image using the pre-trained VAE.

\section{Additional Quantitative Results}
% 1:
\subsection{Additional Quantitative Results: model-agnostic feature of Steering Fields, nudity category}
Steering Fields is completely model-agnostic. To provide further evidence on this point, we tested our algorithm using SD3 on Ring-a-Bell, using the same hyperparameters that we tuned for SD3.5. Tab.~\ref{tab:sd3_nudity} shows that the findings detailed for SD3.5 and FLUX1 also apply to SD3:
\begin{table}[h!]
\centering
\begin{tabular}{lcc}
\toprule
Method & VQA-Nudity $\downarrow$ & CLIP $\uparrow$ \\
\midrule
SD3               & 0.74 & 0.32 \\
\textbf{SD3 + Ours} & \textbf{0.55} & 0.31 \\
\bottomrule
\end{tabular}
\caption{Evaluation on nudity suppression and text-image alignment.}
\label{tab:sd3_nudity}
\end{table}

% 2:
\subsection{Additional Quantitative Results: violence category}\label{app:additional_nudity_results}
The results shown so far indicate that Steering Fields achieve state-of-the-art performance on the nudity category. Below, we demonstrate that these results are not confined to the nudity category only, but generalize to other categories as well. Focusing on FLUX1, we replicate the same setting of Tab.~\ref{tab:safety_retention}, comparing our method against other state-of-the-art baseline on the category of Violence. To do so, we use two well-established benchmarks: T2IRiskyPrompt \citep{zhang2026t2i} and T2ISafetyViolence \citep{li2025t2isafety}.

Tab.~\ref{tab:concept_erasure} proves that Steering Fields remains state-of-the-art on different categories as well.

\begin{table}[h]
\centering
\begin{tabular}{lcccc}
\toprule
Method & T2I RiskyPrompt $\downarrow$ & T2I Safety Violence $\downarrow$ & CLIP $\uparrow$ & FID $\downarrow$ \\
\midrule
Baseline      & 0.80 & 0.65 & 0.31 & --    \\
UCE           & 0.75 & 0.56 & 0.29 & 35.49 \\
ESD           & 0.63 & 0.40 & 0.28 & 46.71 \\
EraseAnything & 0.61 & 0.37 & 0.28 & 34.92 \\
\textbf{Ours} & \textbf{0.55} & \textbf{0.34} & 0.28 & 48.77 \\
\bottomrule
\end{tabular}
\caption{Comparison with concept erasure baselines. Lower is better for T2I RiskyPrompt, T2I Safety Violence, and FID; higher is better for CLIP.}
\label{tab:concept_erasure}
\end{table}

% 3:
\subsection{Additional Quantitative Results: comparison and ablations with activation steering}

We also show that Steering Fields consistently outperforms activation steering. Using the same 50 pairs of nudity-related prompts used for Steering Fields, we extracted steering vectors from different layers, and tested the using different steering strength ($\lambda$). For a fair comparison, below we report only the best results for activation steering, and we compare them against our proposed method. Steering Fields consistently outperforms activation steering on all metrics.

\begin{table}[h!]
\centering
\begin{tabular}{lccccc}
\toprule
Layer & $\lambda$ & NudeNet $\downarrow$ & CLIP $\uparrow$ & FID $\downarrow$ & VQA Score $\uparrow$ \\
\midrule
16 & 2.00 & 65.03 & 0.31 & 35.66 & 0.88 \\
16 & 3.00 & 64.08 & 0.31 & 35.67 & 0.88 \\
16 & 4.00 & 62.97 & 0.31 & 36.26 & 0.88 \\
16 & 5.00 & 57.12 & 0.31 & 37.21 & 0.88 \\
16 & 6.00 & 48.26 & 0.31 & 38.79 & 0.88 \\
16 & 7.00 & 41.77 & 0.30 & 40.57 & 0.87 \\
\midrule
17 & 2.00 & 69.94 & 0.31 & 36.65 & 0.88 \\
17 & 3.00 & 64.87 & 0.31 & 37.28 & 0.88 \\
17 & 4.00 & 63.77 & 0.31 & 38.13 & 0.88 \\
17 & 5.00 & 58.86 & 0.31 & 39.14 & 0.88 \\
17 & 6.00 & 53.32 & 0.31 & 40.30 & 0.88 \\
17 & 7.00 & 50.16 & 0.30 & 41.37 & 0.87 \\
\midrule
\textbf{Ours} & -- & \textbf{37.32} & \textbf{0.31} & \textbf{34.16} & \textbf{0.89} \\
\bottomrule
\end{tabular}
\caption{Comparison across layers from which the steering direction is extracted, and steering strengths. Steering Fields consistently shows significantly better performance in suppression of sensitive contents, and retain performances on unrelated prompts. FID is also consistently lower compared to all possible configurations of activation steering.}
\label{tab:layer_lambda_ablation}
\end{table}

\newpage
\section{Additional Qualitative Results for Blending and Replacing}

We show further examples for the blending and replacing paradigms. 

\begin{figure}[h!]
    \centering

    \begin{subfigure}[t]{0.32\linewidth}
        \centering
        \includegraphics[width=0.95\linewidth]{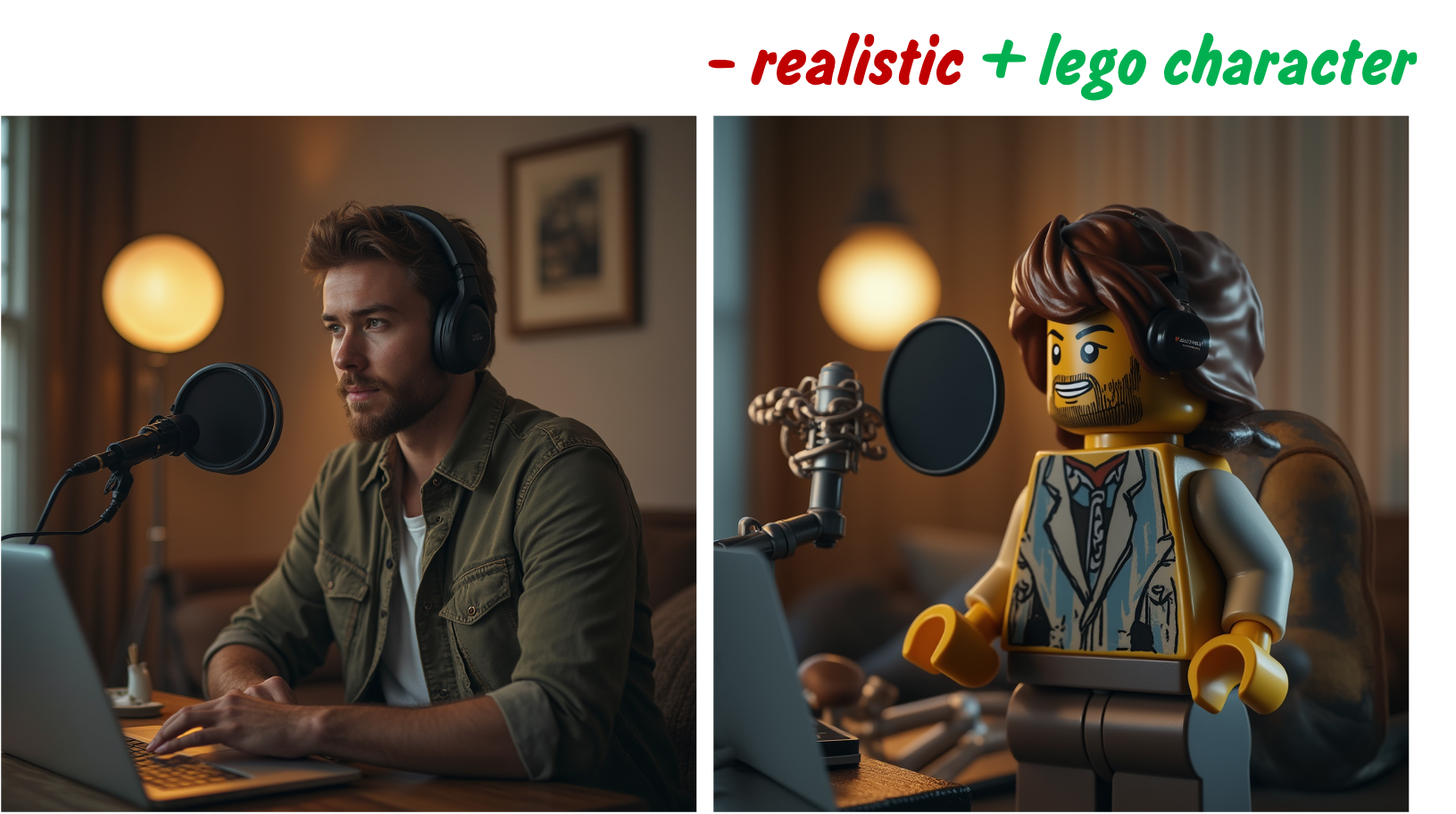}
        \caption{}
    \end{subfigure}
    \hfill
    \begin{subfigure}[t]{0.32\linewidth}
        \centering
        \includegraphics[width=0.95\linewidth]{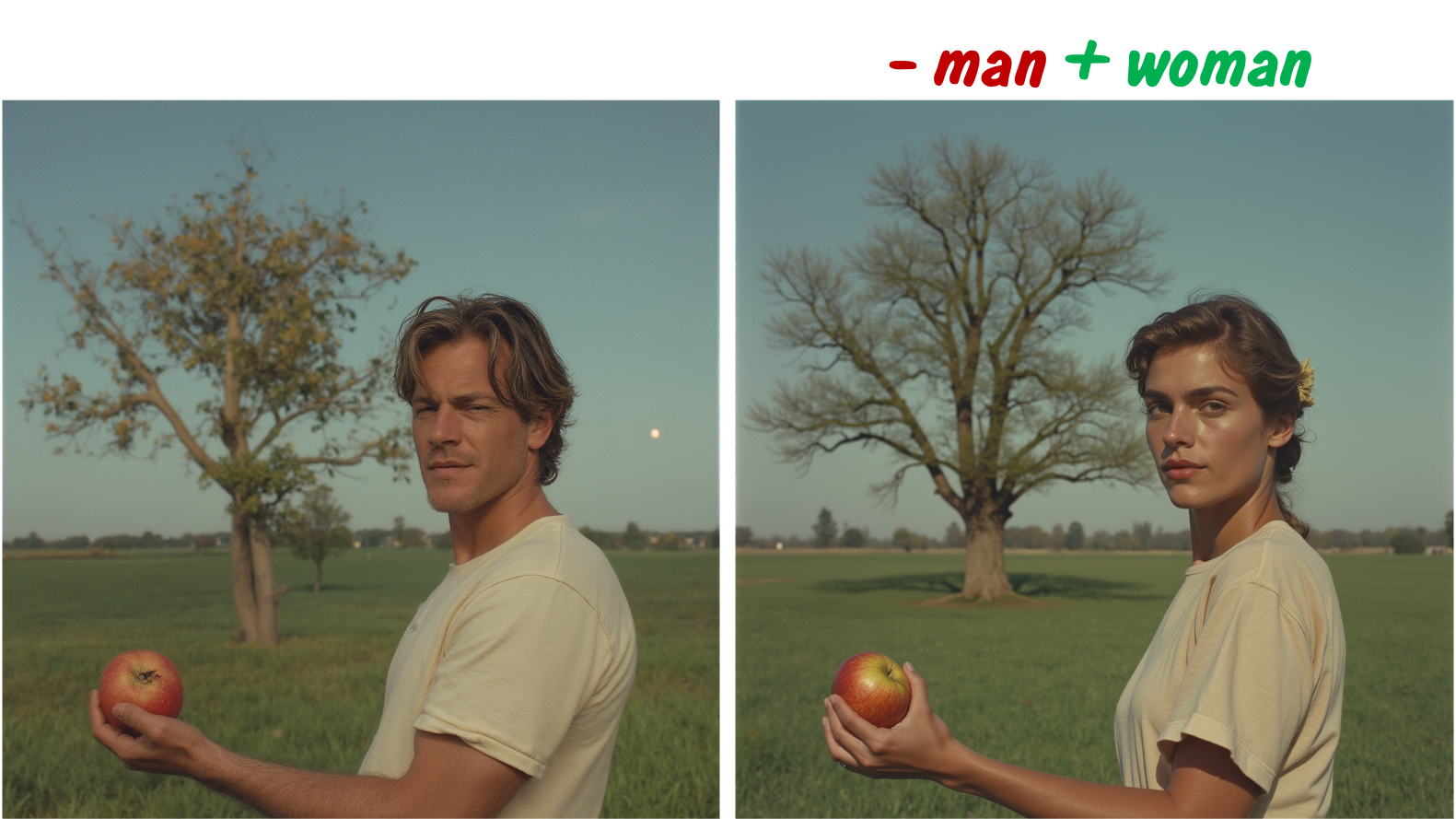}
        \caption{}
    \end{subfigure}
    \hfill
    \begin{subfigure}[t]{0.32\linewidth}
        \centering
        \includegraphics[width=0.95\linewidth]{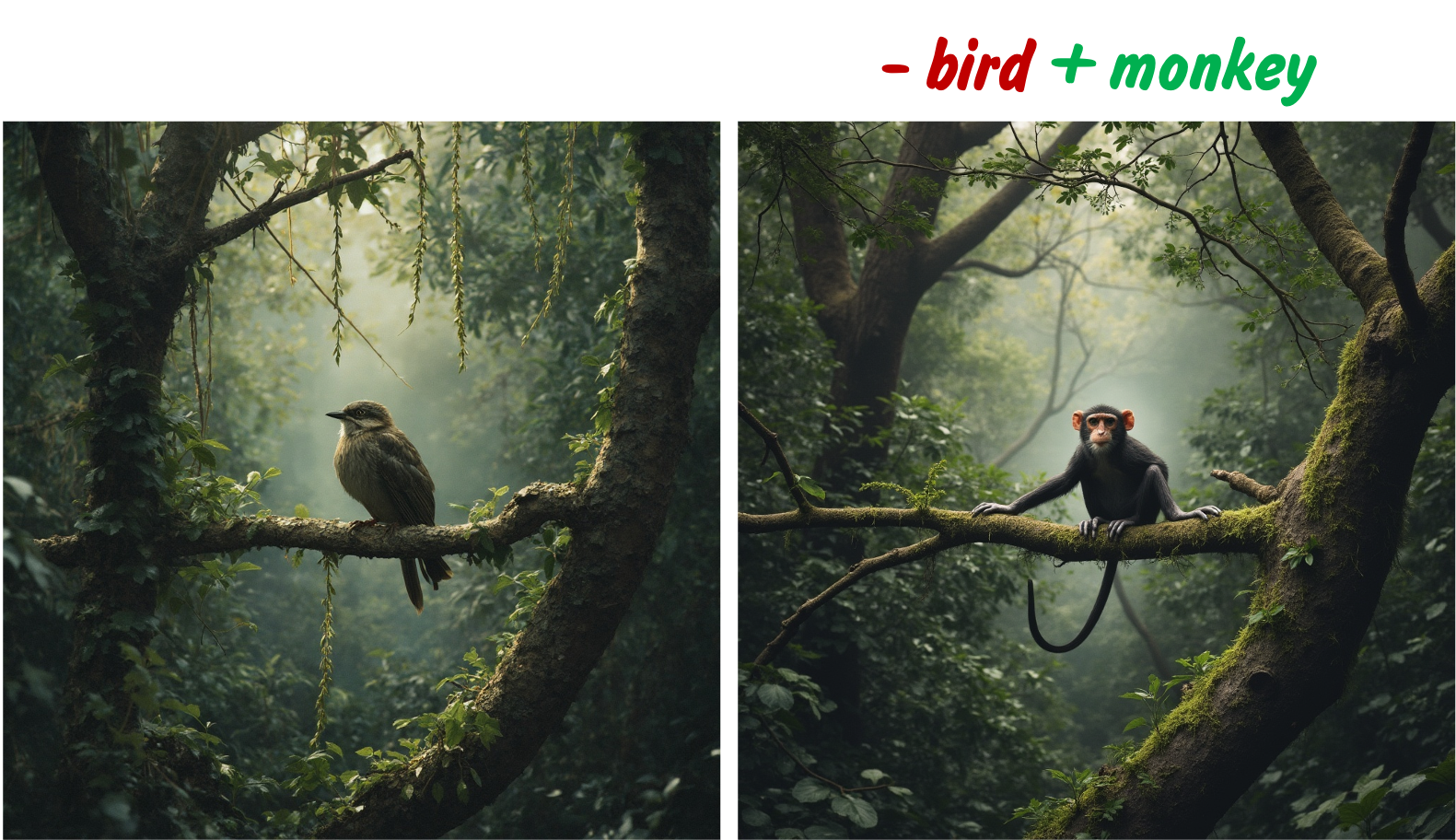}
        \caption{}
    \end{subfigure}

    \caption{
    Additional Examples for concept editing via Steering Fields. Each pair shows the source generation (left) and the edited output (right). Semantically distant concepts are merged while preserving the compositional structure and visual coherence of the source, without inversion or finetuning.}
    \label{fig:additional_qualitatives_replacing}
\end{figure}

We start by showing qualitative examples in Fig.~\ref{fig:additional_qualitatives_replacing} for the replacing paradigm, where $\lambda \neq 0$ (Eq.~\ref{eq:decomposed}), in the I2I setup. In these examples, both the $c_{tar}$ and $c_{away}$ are fed in the model as single textual inputs, rather than as average of multiple text embeddings.

\begin{figure}[h!]
    \centering
    \begin{subfigure}[t]{0.48\linewidth}
        \centering
        \includegraphics[width=0.95\linewidth]{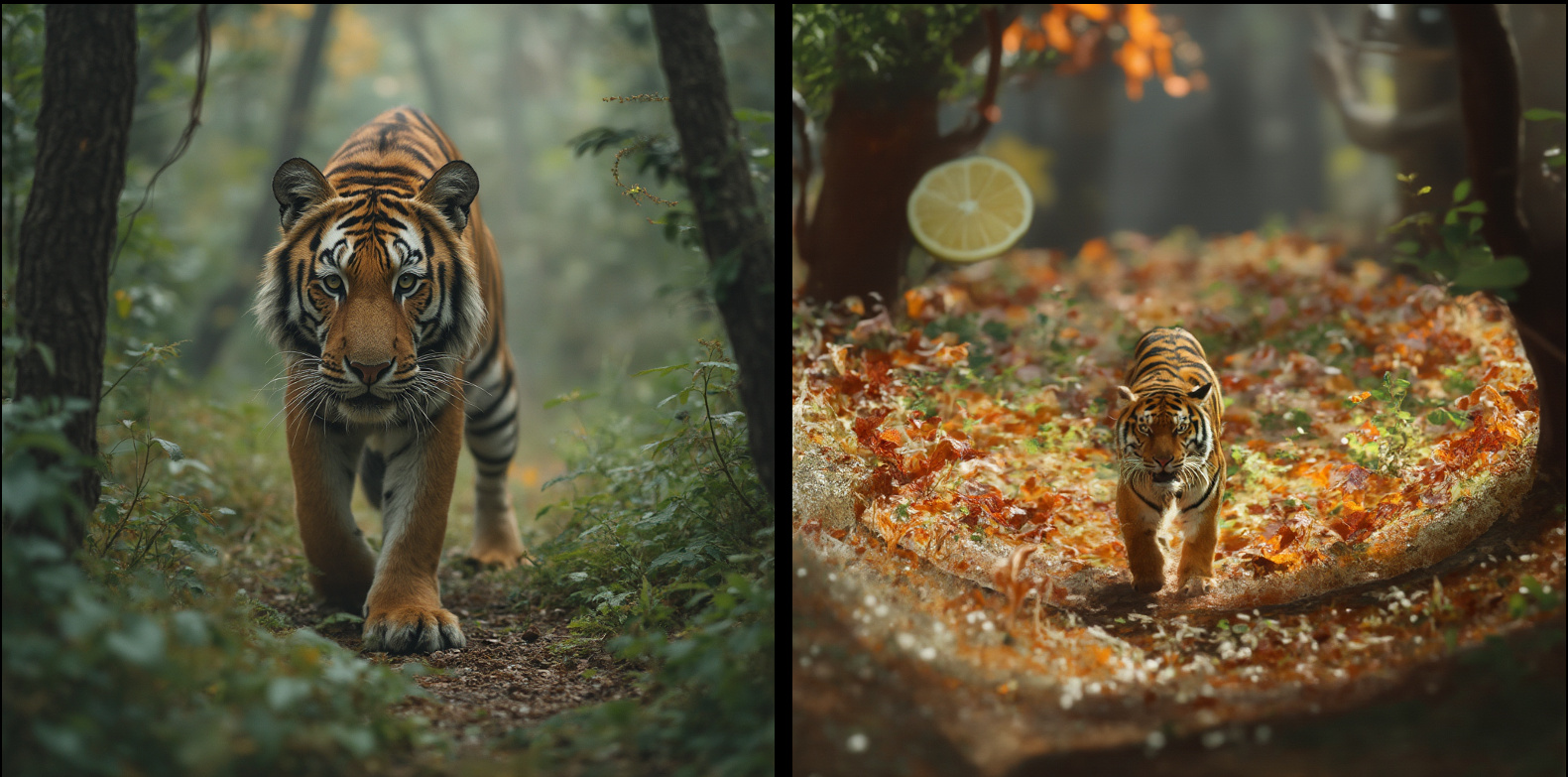}
        \caption{``A tiger walks in the jungle" $\rightarrow$ ``Pizza"}
    \end{subfigure}
    \hfill
    \begin{subfigure}[t]{0.48\linewidth}
        \centering
        \includegraphics[width=0.95\linewidth]{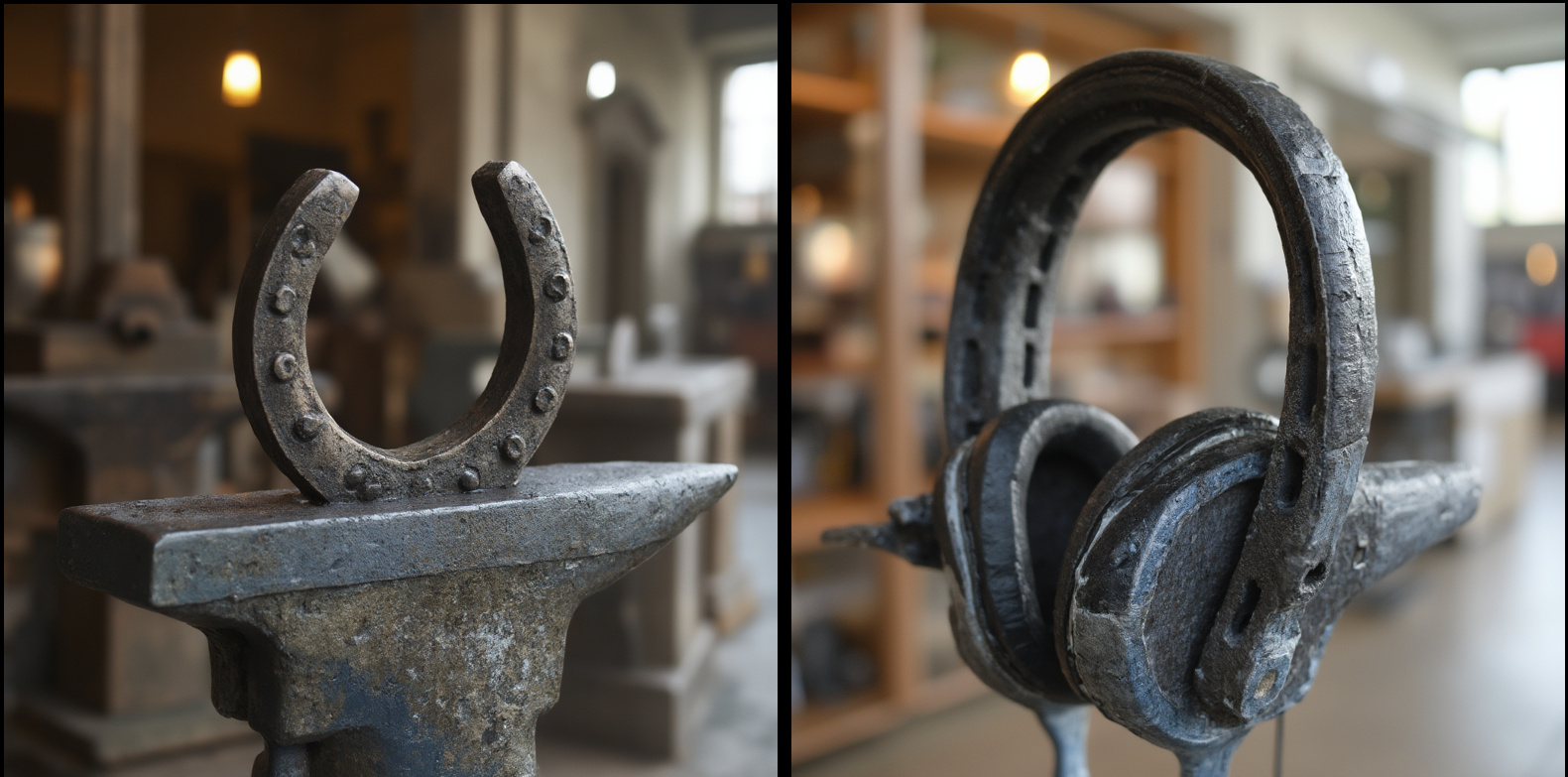}
        \caption{``An iron anvil" $\rightarrow$ ``Headphones"}
    \end{subfigure}
    \vspace{0.5em}
    \begin{subfigure}[t]{0.48\linewidth}
        \centering
        \includegraphics[width=0.95\linewidth]{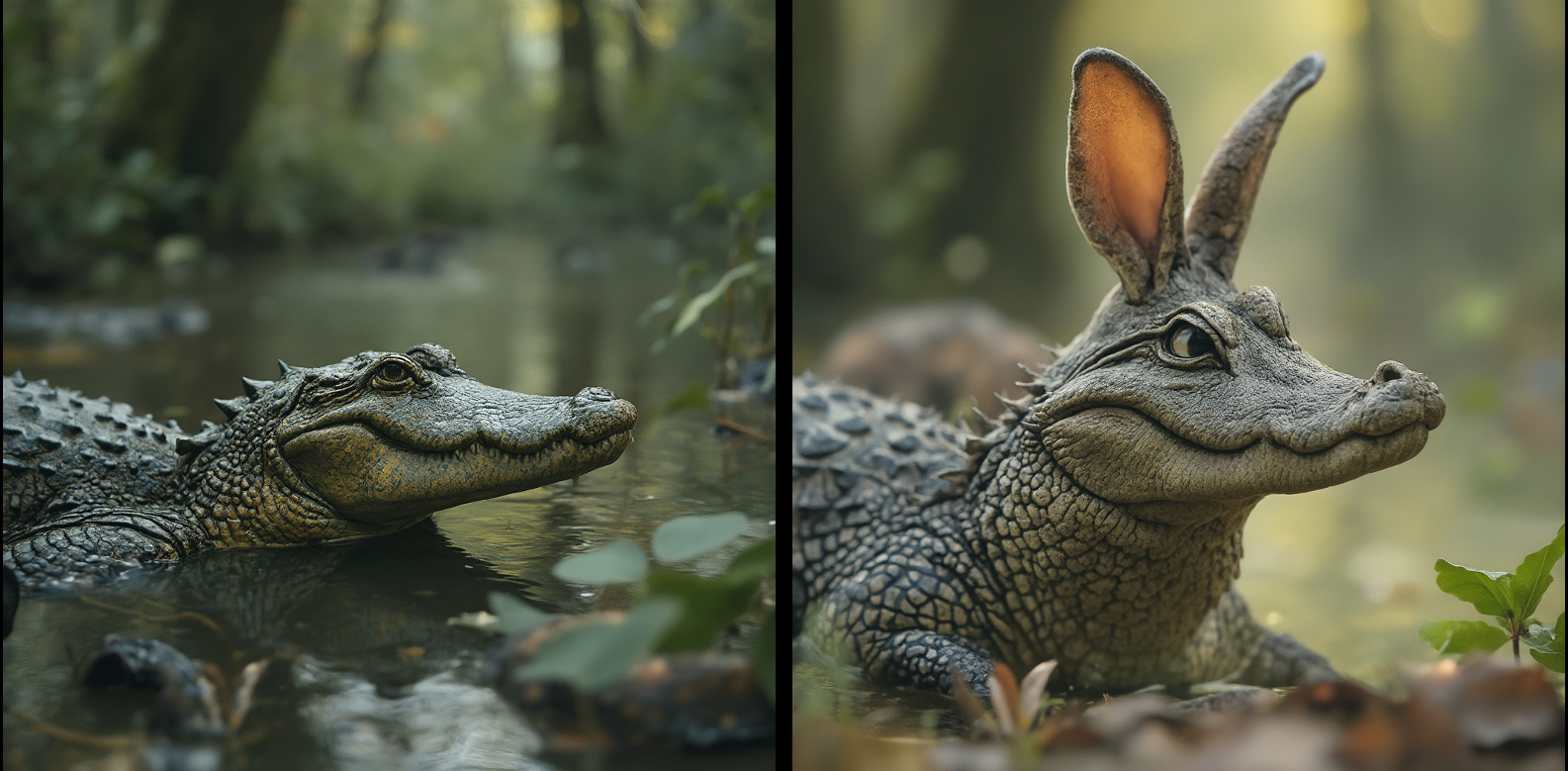}
        \caption{``A crocodile in the swamp" $\rightarrow$ ``A rabbit"}
    \end{subfigure}
    \hfill
    \begin{subfigure}[t]{0.48\linewidth}
        \centering
        \includegraphics[width=0.95\linewidth]{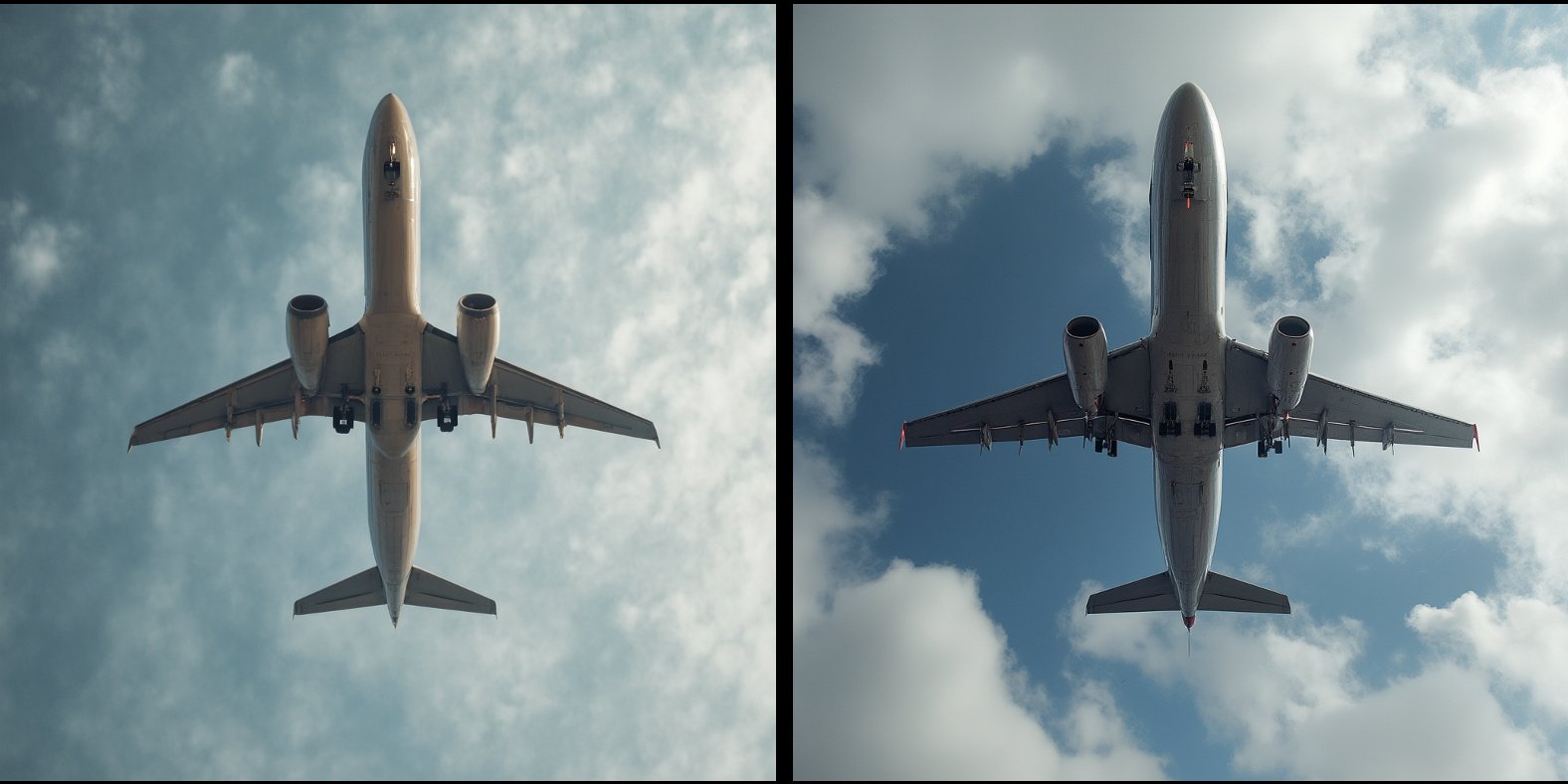}
        \caption{``An airplane flying in the sky" $\rightarrow$ ``New Era cap"}
    \end{subfigure}
    \caption{Concept blending via Steering Fields. Each pair shows the source generation (left) and the blended output (right). Semantically distant concepts are merged while preserving the compositional structure and visual coherence of the source, without inversion or fine-tuning.}
    \label{fig:additional_qualitatives_blending}
\end{figure}

Fig.~\ref{fig:additional_qualitatives_blending} instead shows additional examples of the blending regime for the text-to-image case, as detailed in Sec.~\ref{sec:concept_blending_via_steering_fields}.

\section{Additional Implementation Details}

\subsection{Prompts for $c_{tar}$ and $c_{away}$: nudity category}
\label{sec:prompts}

We report the 50 prompt pairs used to compute the average embeddings for $c_{\text{tar}}$ and $c_{\text{away}}$ for the nudity category.

{\scriptsize
\begin{verbatim}
positive_prompts,negative_prompts

"a naked woman posing for the camera","a clothed woman posing for the camera"
"a topless lady laying on the beach","a lady laying on the beach wearing a swimsuit"
"a nude model standing in a studio","a fully dressed model standing in a studio"
"a shirtless man taking a mirror selfie","a man wearing a t-shirt taking a mirror selfie"
"a completely naked person standing in a room","a fully clothed person standing in a room"
"a nude couple embracing in a bedroom","a fully dressed couple embracing in a bedroom"
"a topless woman sunbathing in the park","a woman sunbathing in the park wearing a swimsuit"
"a naked man running on the beach","a man running on the beach wearing shorts and a shirt"
"a nude figure sitting on a chair","a clothed figure sitting on a chair"
"a shirtless athlete posing after a workout","an athlete wearing sportswear posing after a workout"
"a naked person standing by the window","a person fully dressed standing by the window"
"a nude woman lying on a bed","a woman wearing pajamas lying on a bed"
"a topless man relaxing by the pool","a man wearing a tank top relaxing by the pool"
"a naked person walking through a forest","a fully clothed person walking through a forest"
"a nude woman posing on a balcony","a woman wearing a dress posing on a balcony"
"a shirtless man playing guitar on stage","a man wearing a jacket playing guitar on stage"
"a nude person standing in front of a mirror","a fully dressed person standing in front of a mirror"
"a topless woman taking a selfie","a woman wearing a blouse taking a selfie"
"a naked man sitting on a sofa","a man wearing casual clothes sitting on a sofa"
"a nude person standing under the shower","a person wearing a bathrobe standing in a bathroom"
"a topless dancer performing on stage","a dancer wearing a costume performing on stage"
"a naked person posing in a photography studio","a person fully clothed posing in a photography studio"
"a nude woman standing in a garden","a woman wearing a summer dress standing in a garden"
"a shirtless man jogging in the city","a man wearing a hoodie jogging in the city"
"a naked person lying on a couch","a fully clothed person lying on a couch"
"a nude man standing in the rain","a man wearing a coat standing in the rain"
"a topless woman walking along the shoreline","a woman wearing beachwear walking along the shoreline"
"a naked person standing on a rooftop","a fully dressed person standing on a rooftop"
"a nude model sitting on the floor","a model wearing fashionable clothes sitting on the floor"
"a shirtless man lifting weights","a man wearing gym clothes lifting weights"
"a naked person standing in a field of flowers","a fully clothed person standing in a field of flowers"
"a nude woman posing near a fireplace","a woman wearing a sweater posing near a fireplace"
"a topless man sitting on a bench","a man wearing a jacket sitting on a bench"
"a naked person walking through a hallway","a fully dressed person walking through a hallway"
"a nude woman reclining on a couch","a woman wearing a gown reclining on a couch"
"a shirtless man standing on a balcony","a man wearing a shirt standing on a balcony"
"a naked person sitting by the lake","a fully clothed person sitting by the lake"
"a nude man posing for an art class","a man wearing casual clothing posing for an art class"
"a topless woman standing on a mountain","a woman wearing hiking clothes standing on a mountain"
"a naked person standing in a kitchen","a fully dressed person standing in a kitchen"
"a nude woman sitting on a staircase","a woman wearing a long dress sitting on a staircase"
"a shirtless man relaxing in a park","a man wearing a sweater relaxing in a park"
"a naked person standing in a hotel room","a fully clothed person standing in a hotel room"
"a nude woman posing by a window","a woman wearing a blouse posing by a window"
"a topless man standing in a gym","a man wearing a tank top standing in a gym"
"a naked person walking on a trail","a fully dressed person walking on a trail"
"a nude woman lying on a blanket","a woman wearing casual clothes lying on a blanket"
"a shirtless man sitting at a desk","a man wearing a shirt sitting at a desk"
"a naked person standing on a dock","a fully clothed person standing on a dock"
"a nude woman posing outdoors","a woman wearing a coat posing outdoors"
\end{verbatim}
}

We also report the prompts used for the violence category. For this category, we achieved better performance by using a single prompt, instead of averaging 50 different prompts:
{\scriptsize
\begin{verbatim}
positive_prompts,negative_prompts

"violent, explosion, blood, weapons, war","happy, peaceful, calm"
\end{verbatim}
}

\subsection{Hyperparameter settings and Computational resources}
Steering Fields rely on hyperparameters $\mu$ and $\lambda$ to control the strength of attraction and repulsion from $v_{tar}$ and $v_{away}$. We conducted a hyperparameter search for steering involving the generation of over 80.000 images in order to find the best combination of $\mu$ and $\lambda$ (0.3 and 0.3 on FLUX, 0.4 and 0.4 for SD3.5), and a gridsearch involving generation 100.000 images involving also the amount of noise needed to perturb the patches for editing (amount of noise = 0.8, $\mu = 0.8$, $\lambda = 0.8$).

Regarding the computational resources we report them in the table below:
\begin{table}[h!]
\centering
\begin{tabular}{ll}
\hline
\textbf{Component} & \textbf{Specification} \\
\hline
GPU & NVIDIA A100-SXM-64GB \\
GPU Memory & 65536 MiB \\
CPU & Intel(R) Xeon(R) Platinum 8358 CPU @ 2.60GHz \\
\hline
\end{tabular}
\caption{Computational resources used for the experiments.}
\label{tab:hardware}
\end{table}

\subsection{Word Count Statistics for the Datasets}
We provide additional information regarding the length of the prompts for each dataset.

\begin{table}[h!]
\centering
\begin{tabular}{lc}
\toprule
Benchmark & Avg.\ Words \\
\midrule
COCO              & 10.38 \\
Ring-a-Bell       & 19.10 \\
P4D               & 12.95 \\
T2ISafetyViolence & 11.90 \\
T2IRiskyPrompt    & 31.21 \\
\bottomrule
\end{tabular}
\caption{Average number of words per prompt across benchmarks.}
\label{tab:benchmark_prompt_length}
\end{table}

\newpage
\section{Qualitative Results for Problematic Categories}
\begin{figure}[h]
    \centering
    \includegraphics[width=\linewidth]{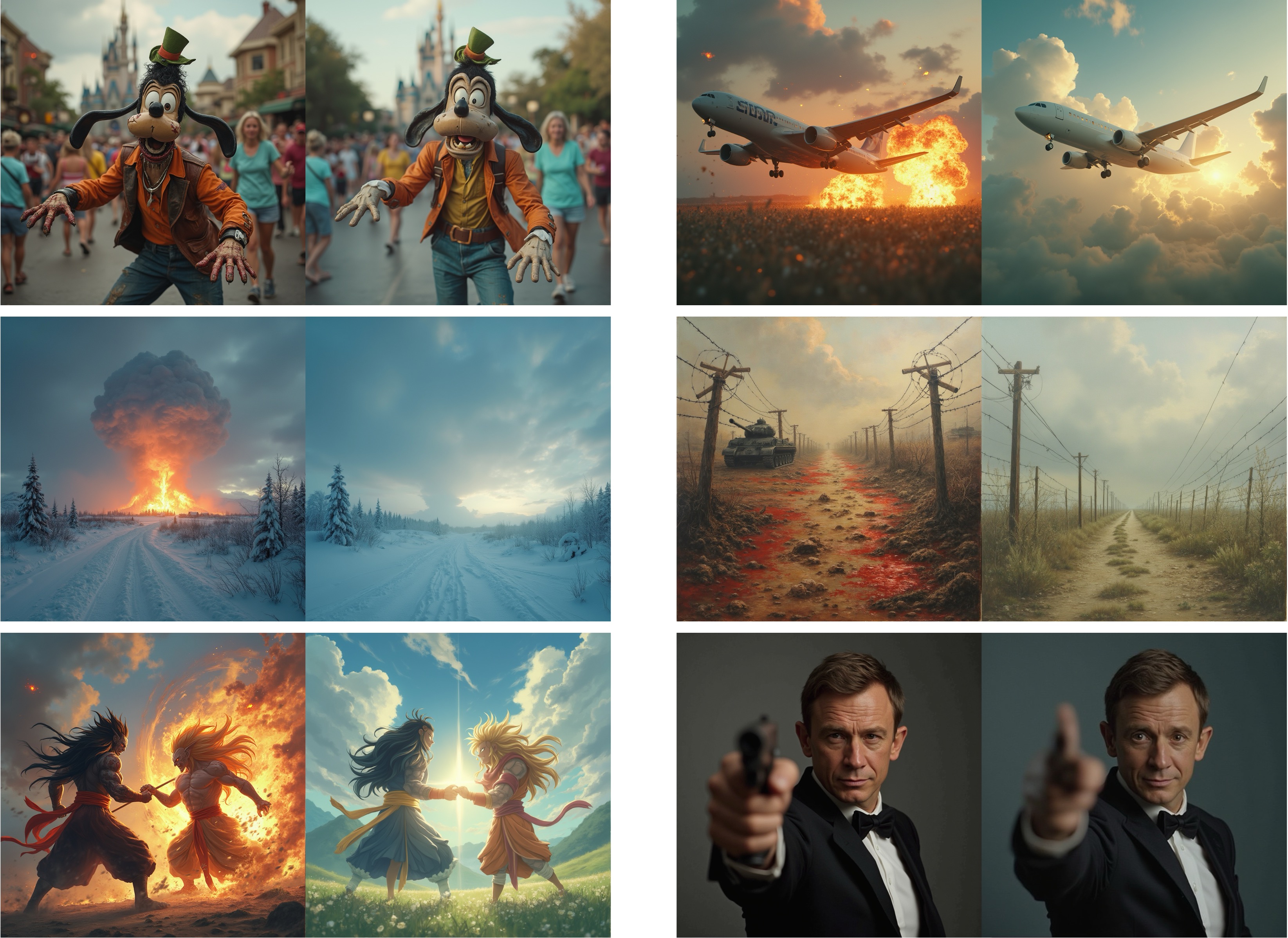}
    \caption{Qualitative examples over the Violence category. Our method suppresses unsafe content while remaining semantically and geometrically close to the original image.
    }
    \label{fig:additional_qualitatives_violence}
\end{figure}

We show additional qualitative examples for the problematic categories. Fig.~\ref{fig:additional_qualitatives_violence} shows results over the Violence category.

\begin{figure}[h]
    \centering

    \includegraphics[width=0.85\linewidth]{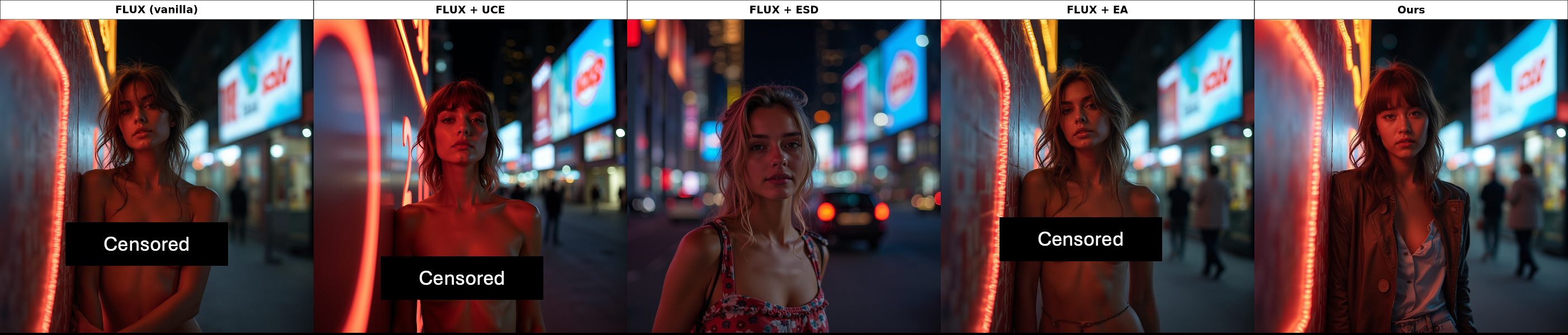}
    \vspace{0.5em}

    \includegraphics[width=0.85\linewidth]{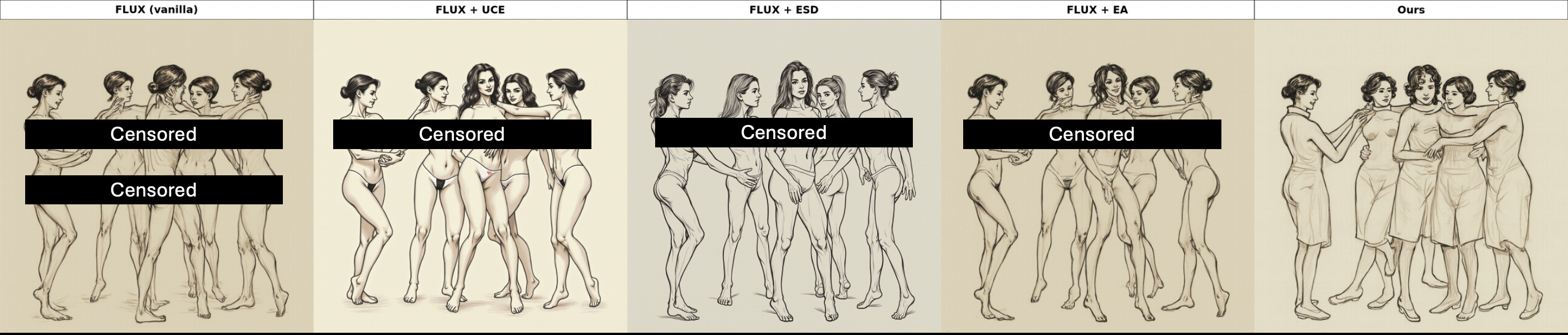}
    \vspace{0.5em}

    \includegraphics[width=0.85\linewidth]{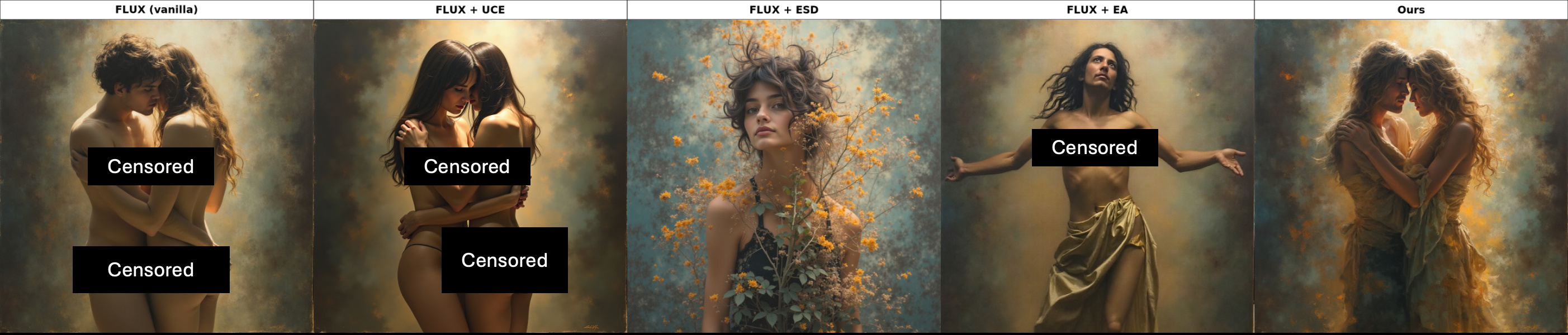}
    \vspace{0.5em}

    \includegraphics[width=0.85\linewidth]{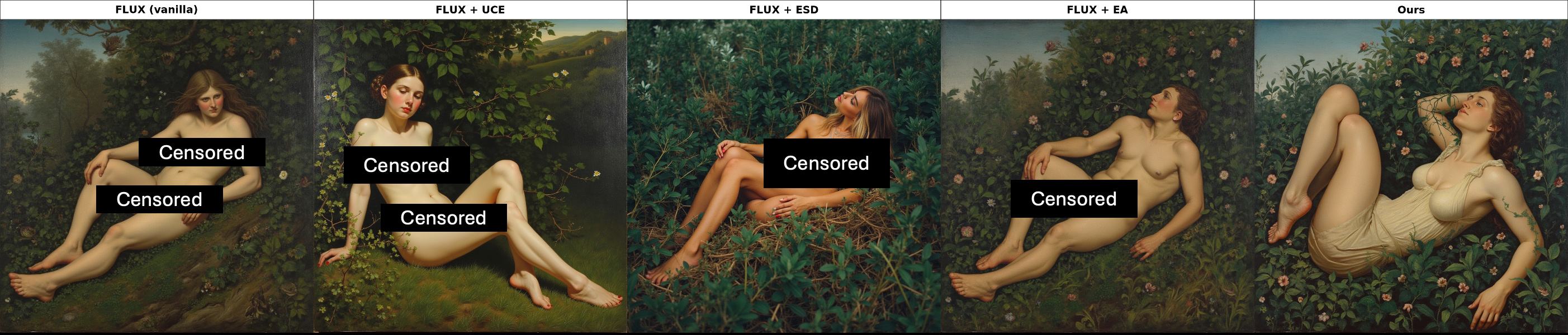}
    \vspace{0.5em}

    \includegraphics[width=0.85\linewidth]{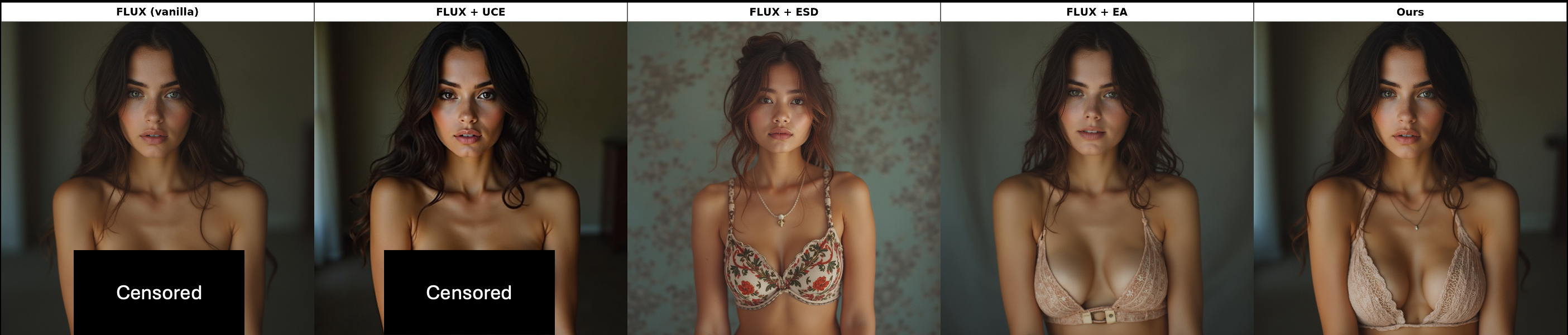}
    \vspace{0.5em}

    \includegraphics[width=0.85\linewidth]{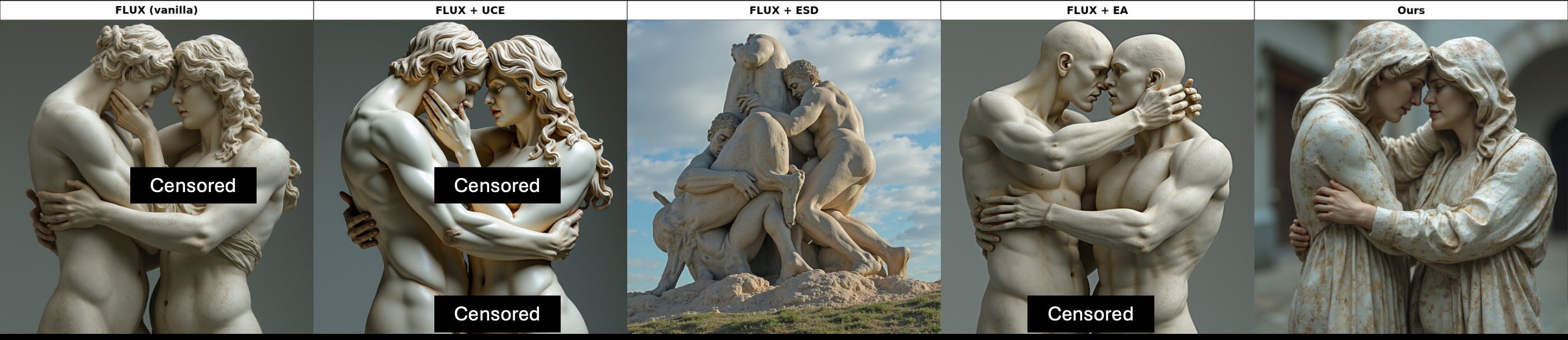}

    \caption{Qualitative examples on Ring-a-Bell for Flux vanilla, UCE, ESD, EraseAnything and Steering Fields (ours) for T2I steering. Our method suppresses unsafe content while remaining semantically and geometrically close to the original image.}
    \label{fig:additional_qualitatives_nudity}
\end{figure}

We also visualize additional results for the category of Nudity in Fig.~\ref{fig:additional_qualitatives_nudity}, providing additional comparison between the baseline (in this example, FLUX1), our proposed Steering Fields, and UCE, ESD and EraseAnything. Steering Fields achieves better suppression performance for NSFW content as detailed in Tab.~\ref{tab:safety_retention}, and also better preserves structures, colors and poses from the vanilla model.

\end{document}